\documentclass[10pt,twocolumn,letterpaper]{article}

\usepackage[pagenumbers]{cvpr}

\usepackage[utf8]{inputenc}
\usepackage[T1]{fontenc}
\usepackage{multirow}
\usepackage{microtype}
\usepackage{comment}

\definecolor{cvprblue}{rgb}{0.21,0.49,0.74}
\usepackage[pagebackref,breaklinks,colorlinks,allcolors=cvprblue]{hyperref}

\def\subsubsubsection{\vskip 5pt{\noindent\normalsize\rm\raggedright}}
\newcommand*\samethanks[1][\value{footnote}]{\footnotemark[#1]}

\title{LiteGUI: Lightweight GUI Agents via Multi-solution Guided Distillation and Offline Dual-level Policy Optimization}

\author{
Yubin Wu\thanks{Equal contribution.}\quad
Zicheng Cai\samethanks{}\quad
Liping Ning\quad
Hua Wang\\
Zhi Chen\quad
Yaohua Tang\thanks{Corresponding author: \texttt{tangyaohua28@gmail.com}.}\quad
Hao Chen\\[1pt]
{\normalsize Moore Threads AI}\vspace{-6pt}
}

\hypersetup{
  pdftitle={LiteGUI: Lightweight GUI Agents via Multi-solution Guided Distillation and Offline Dual-level Policy Optimization},
  pdfauthor={Yubin Wu, Zicheng Cai, Liping Ning, Hua Wang, Zhi Chen, Yaohua Tang, Hao Chen}
}

\begin{document}

\maketitle

\begin{abstract}

We present LiteGUI, a post-training framework for building lightweight vision-only GUI agents. Visual GUI interaction is challenging for compact models because desktop interfaces expose multiple equivalent action affordances (e.g., context menus, toolbar icons, and keyboard shortcuts) and single-reference imitation quickly collapses into repetitive action loops once execution deviates from a fixed template. LiteGUI addresses these challenges through two-stage optimization over logged GUI states without requiring costly online environment rollouts during training. First, we propose \textit{Guided Multi-Solution On-Policy Distillation (Guided-OPD)}, where the student samples completions from its current policy at logged visual states and receives privileged teacher guidance conditioned on the human-verified valid action most matched to the student's sampled behavior. Second, we develop \textit{Multi-Solution Dual-Level GRPO (MD-GRPO)}, which combines set-valued visual action matching with history-conditioned subtask-plan quality supervision to ground step-by-step progress in screenshot evidence. We further develop a scalable data generation pipeline and a corresponding multi-solution dataset, \textit{Lite-Dataset}, to support the training and evaluation of GUI agents, addressing a gap in the existing literature. Deployed in live closed-loop GUI environments with zero inference-time judge overhead, LiteGUI-2B more than doubles OSWorld task success over its backbone (from 6.04\% to 13.24\%), while LiteGUI-30B-A3B (3B active parameters) achieves 58.95\% on ScreenSpot-Pro, 22.7\% on OSWorld, and 88.75\% on Lite-Bench.

\end{abstract}

\section{Introduction}
Automated computer use has received increasing attention in recent years \citep{hurst2024gpt,manus,claude,team2023gemini,Qwen-VL,Qwen3-VL}, with vision-only GUI agents interpreting screenshots to execute desktop workflows. While prior systems rely predominantly on large-scale models \citep{qin2025ui,liu2025pc,agashe2025agent}, building lightweight models\footnote{By a lightweight model, we refer to models in the 2--3B active-parameter range.} for low-latency deployment remains challenging. In multi-step visual interfaces, standard supervised fine-tuning (SFT) forces compact models to imitate a single logged demonstration path: when the agent encounters equivalent UI affordances (e.g., right-clicking, clicking a toolbar icon, or pressing a hotkey) or minor state shifts, single-path imitation penalizes valid alternatives and frequently degrades into unverified subtask hallucinations and repetitive click loops.

A natural default for training GUI agents is online interaction: build an environment, let the agent act, and optimize from the resulting feedback. This setting, however, requires repeated environment resets, safe execution, and reproducible rollouts, which makes systematic training costly. At the same time, logged GUI trajectories are abundant but underused: existing offline methods usually reduce them to single-action imitation targets or single-answer rewards, discarding the multiple valid actions that make GUI data valuable. Offline optimization is therefore a deliberate choice, not a limitation: we start from logged GUI states, recover this multi-solution signal, and use closed-loop evaluation to test whether the resulting policy transfers to real execution. In this paper, we propose \textbf{LiteGUI}, an offline post-training framework for lightweight GUI agents, built on this idea.

First, we propose \textbf{Guided On-Policy Distillation (Guided-OPD)}, where the student samples completions from its current policy at logged visual states rather than executing actions online. Because open-ended teacher estimates can be noisy on complex screens, Guided-OPD conditions the teacher on training-time privileged information selected from human-verified valid-action sets $\mathcal{A}_t^*$. Specifically, \textbf{Most-Matched valid-action selection} aligns teacher guidance with the valid action closest to the student's sampled mode, raising teacher pairwise preference accuracy from 46\% to 77\% across 50 validation states without restricting student exploration to a single template.

Second, we develop \textbf{Multi-Solution Dual-Level GRPO (MD-GRPO)}, which couples set-valued micro-level action matching with a macro-level, history-conditioned subtask-plan quality signal scored on the three-step screenshot window. Together, these local signals reward any valid interaction modality while grounding subtask state transitions in visual evidence. In summary, our main contributions are:
\begin{itemize}
\item \textbf{LiteGUI Post-Training Framework:} We introduce a two-stage logged-state optimization paradigm for lightweight vision-only GUI agents, combining \textbf{Most-Matched Guided-OPD} (adaptive privileged teacher guidance over valid action sets) with \textbf{MD-GRPO} (dual-level set-valued action matching and history-conditioned subtask planning).
\item \textbf{Multi-Solution Data Pipeline and Benchmark:} We build a semi-automated trajectory exploration pipeline with human verification and release \textbf{Lite-Dataset} (30K trajectories, $\sim$70K windowed states, and 11K multi-solution annotations) alongside \textbf{Lite-Bench} (160 live Ubuntu tasks across File System, Web, and Terminal).
\item \textbf{Strong Closed-Loop Transfer at 2B--3B Active Scale:} Across ScreenSpot-Pro \citep{li2025screenspotpro}, OSWorld \citep{OSWorld}, and Lite-Bench, LiteGUI-2B more than doubles OSWorld success over its backbone (6.04\% to 13.24\%) and reaches 63.13\% on Lite-Bench, while LiteGUI-30B-A3B (3B active parameters) reaches 58.95\% on ScreenSpot-Pro, 22.7\% on OSWorld, and 88.75\% on Lite-Bench with zero inference-time judge overhead.
\end{itemize}

\begin{figure*}[t]
  \centering
    \includegraphics[width=\textwidth]{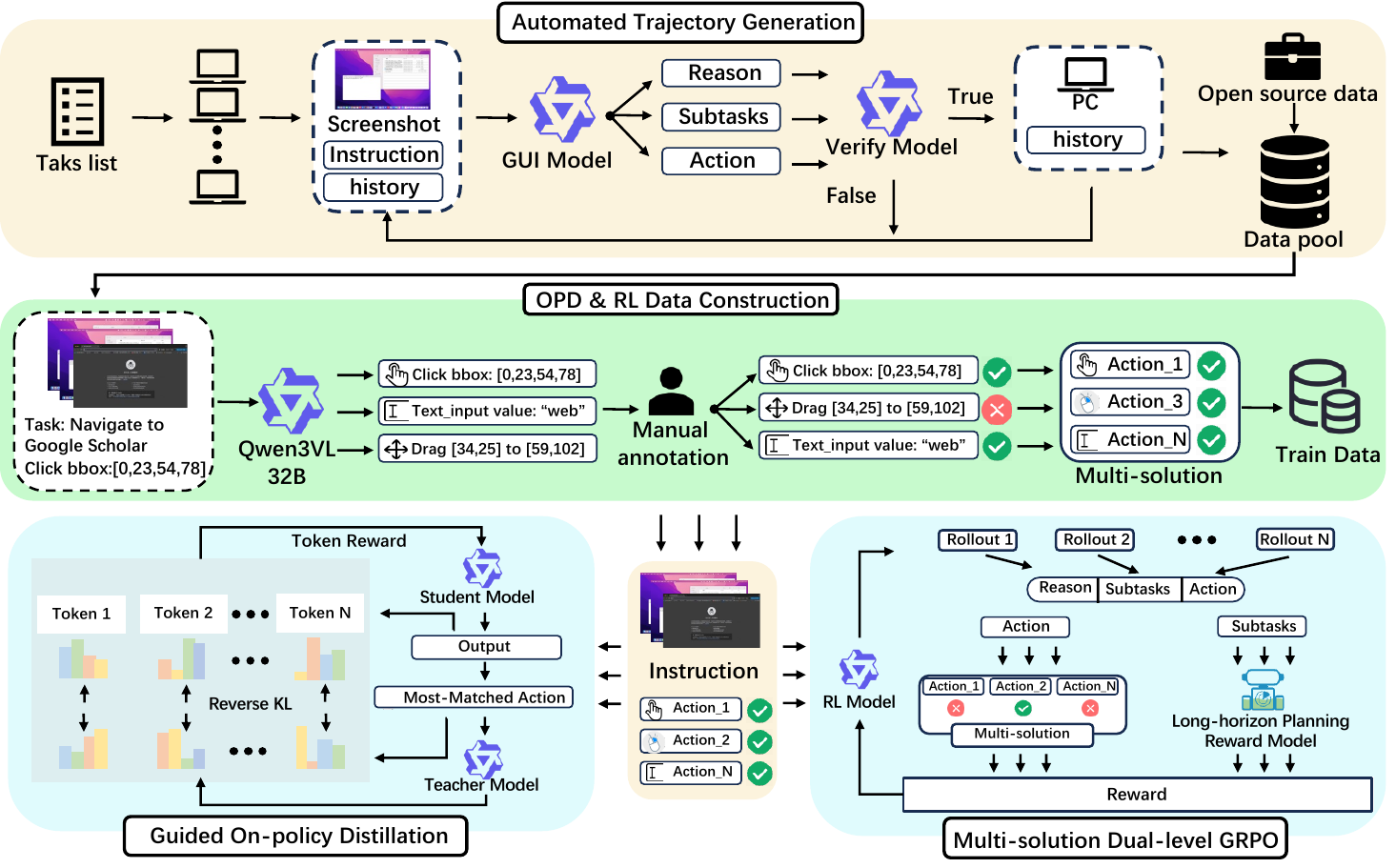}
  \caption{Overview of LiteGUI. The framework consists of: (a) an automated GUI trajectory generation module for producing action trajectories for GUI tasks; (b) an OPD and policy-optimization data construction module for generating training data, with a particular focus on multi-solution annotations for OPD and policy optimization; and (c) the proposed two-stage training paradigm, comprising Guided On-policy Distillation and Multi-solution Dual-level GRPO. Details of the framework can refer to Appendix~\ref{app:data_pipeline}.}
  \label{fig:acc}
\end{figure*}

\section{Related Work}
\subsection{Vision-based GUI Agents}
Early GUI agents predominantly relied on underlying system APIs or accessibility trees, which severely hindered their cross-platform generalization \citep{deng2023mind2web,OSWorld}. Recently, the research paradigm has rapidly shifted toward vision-only agents, primarily bifurcating into two approaches: the direct application of powerful general-purpose Multimodal Large Language Models (MLLMs) (e.g., GPT-4o\citep{hurst2024gpt}, Claude 3.5 Sonnet, Qwen-VL\citep{Qwen-VL,Qwen3-VL}), and the fine-tuning of domain-specific GUI vision models (e.g., CogAgent \citep{hong2024cogagent}, UI-TARS \citep{qin2025ui}, OpenCUA \citep{wang2025opencuaopenfoundationscomputeruse} and others \citep{lin2024showui,yuan2025enhancing}) utilizing large-scale interaction data \citep{gou2025uground,wu2024atlas,xu2025deskvision}. While these models exhibit exceptional planning and execution performance in the GUI domain, their formidable capabilities incur prohibitive computational overhead during inference, severely impeding practical deployment and user experience. Consequently, developing lightweight agents (e.g., 2B/A3B-scale models) has become an urgent imperative to address these bottlenecks. However, the post-training of current lightweight models relies almost exclusively on traditional Supervised Fine-Tuning (SFT) \citep{hsieh2025zonui3b,xu2025deskvision} or Reinforcement Learning (RL) \citep{shen2025vlm,luo2025gui} methods utilizing naive coordinate rewards (e.g., GRPO). This coarse-grained training paradigm is precisely the root cause of the policy rigidity and capability degradation observed in small-scale models during long-horizon, multi-step exploration.
\subsection{Post-training for GUI Agents}
To improve the interaction capabilities of GUI agents, post-training paradigms are shifting from static imitation toward dynamic exploration. However, most existing approaches still rely on SFT over static trajectories, which suffers from covariate shift in multi-step decision-making \citep{ross2011reduction}, leading to rapid error accumulation once deviating from expert paths. Moreover, fitting long sequential data can disrupt the visual representation capabilities of lightweight models, resulting in catastrophic forgetting \citep{zeng2024agenttuning}. To address these limitations, recent work explores policy distillation methods (e.g., GKD \citep{agarwal2024onpolicy}, SDFT \citep{shenfeld2026self}, and others \citep{chen2510retaining}), where teacher models provide real-time guidance during student exploration. However, such approaches remain underexplored in GUI agents, primarily due to the susceptibility of large models to hallucinations in “closed-book” settings \citep{alansari2026large,huang2025survey}, which introduce noisy supervision and degrade multi-step trajectory execution. We therefore propose Guided On-policy Distillation 
to leverage human-verified valid-action annotations to mitigate hallucinations and employ most-matched valid-action selection to accommodate the multi-solution nature of GUI tasks, enabling flexible action-level exploration.
Furthermore, reinforcement learning (RL) methods (e.g., GRPO \citep{shao2024deepseekmathpushinglimitsmathematical} and SAPO \citep{gao2025softadaptivepolicyoptimization}) have been applied to vision-based models (e.g., VLM-R1 \citep{shen2025vlm}). However, existing reward designs based on exact coordinate/action matching \citep{lu2025ui,luo2025gui} suffer from sparsity and false negatives, limiting generalization. To address this, we introduce a multi-solution dual-level logged-state policy-optimization mechanism: macro-level subtask-plan quality evaluated by a strong model, and micro-level matching over multi-solution actions. These signals are computed on logged GUI states and provide local training feedback without executing candidate actions during optimization.

\section{Method}
\label{sec:method}

We propose a two-stage training paradigm for GUI agents, shown in Fig.\ref{fig:acc}. In the first stage, we introduce \textbf{Guided On-policy Distillation (Guided-OPD)}, which leverages multi-solution action annotations as training-time privileged information to provide a more stable supervisory signal for distillation. In the second stage, we introduce \textbf{Multi-solution Dual-level GRPO (MD-GRPO)}, which uses a multi-solution action signal and a history-conditioned, per-state subtask-plan signal. Both signals are computed on logged GUI states; candidate outputs are not executed during optimization, so they should be interpreted as offline surrogate supervision rather than environment returns. The trained policy is evaluated separately through closed-loop execution in OSWorld and Lite-Bench. To support this protocol, we develop a data generation pipeline to semi-automatically produce multi-solution GUI task trajectories for training. Details of the pipeline are illustrated in Fig.~\ref{fig:acc}. In the following subsections, we first introduce the task formulation of the GUI agent, followed by the details of Guided-OPD and MD-GRPO.

\subsection{Task Formulation}
\label{sec:task_formulation}

Given a user instruction $q$ (e.g., lower the computer volume), the GUI model input $x_t$ is denoted as: 
\begin{equation}
x_t^{(w)} =
\left(q, s_{t-w}, y_{t-w}, \dots, s_{t-1}, y_{t-1}, s_t \right),
\end{equation}
where $s_t$ denotes the GUI screenshot at time step $t$, $y_t$ denotes the structured model output. $w$ is the observation window size, i.e., the input consists of the user instruction, the current state, and the previous states and outputs within a history window of size $w$. The model output $y_t$ contains three key components: 
\begin{equation}
y_t = \{r_t, c_t, a_t\},
\end{equation}
where $r_t$ denotes the \texttt{Reasoning} field, $c_t$ denotes the \texttt{subtask} list, and $a_t$ denotes the executable GUI action (e.g., the mouse and keyboard actions). The \texttt{Reasoning} field follows an Observation--Intent--Analysis structure. For example, it may describe the current UI observation, infer that the user intends to open a website, and conclude that the next step should be clicking the address bar. The \texttt{subtask} field serves as an explicit planning and memory state for the agent, provides a compact mechanism for carrying long-horizon task state across steps. For example, it may contain items such as ``enter the target URL'', ``log into the account'', and ``check whether the homepage has loaded''. These entries store task-level information that may not be fully recoverable from the latest screenshot alone, including subtask completion status, key intermediate milestones, and visual clues extracted from previous observations. Thus, the recent screenshots provide local perceptual context, while subtasks preserve higher-level progress information.
We represent the executable action as
\begin{equation}
a_t = (\tau_t, p_t, v_t),
\end{equation}
where $\tau_t$ is the action type, $p_t$ is the position argument, and $v_t$ is the text or key-value argument. For example, a click action can be represented as $\tau_t=\texttt{CLICK}$ with $p_t=(521,41)$, while a text-input action can be represented as $\tau_t=\texttt{TEXT\_INPUT}$ with $v_t=\texttt{www.example.com}$. The model predicts a point $p_t=(x,y)$; for grounding supervision, the corresponding human annotation may specify a target bounding box $b^*=[x_{\min},y_{\min},x_{\max},y_{\max}]$.

For each state $x_t$, we maintain a set of human-verified valid actions as ground truths:
\begin{equation}
\mathcal{A}^{*}_t = \{a^{*(1)}_t, a^{*(2)}_t, \dots, a^{*(K)}_t\},
\end{equation}
which contains $K$ multiple actions that can correctly advance the task from the current GUI state. It should be viewed as a finite approximation of the valid action space rather than an exhaustive set of all possible correct actions. Because logged prefixes $(s_{t-w:t}, y_{t-w:t-1})$ act as observed conditioning states rather than re-rolled rollouts, scoring alternative valid actions in $\mathcal{A}_t^*$ or updated plans $c_t$ at step $t$ needs no counterfactual future frames; at deployment, live screenshots and self-generated subtask states dynamically replace logged history.

\subsection{Guided On-policy Distillation}

In standard Generalized Knowledge Distillation (GKD) or On-policy Distillation (OPD), the student first samples an output $\hat{a}_t$ conditioned on the current input $x_t$, and the teacher evaluates the sampled sequence under teacher forcing. 
This evaluation produces full next-token distributions at all valid completion prefixes. However, in GUI agent training, student outputs are often noisy: they may contain incorrect action types, inaccurate coordinates, wrong text arguments, or invalid JSON structures. If the teacher evaluates such outputs using only the default context, the resulting likelihood signal can be unstable. To mitigate this issue, we introduce a Guided On-policy Distillation (Guided OPD) mechanism, which provides the teacher with additional action guidance constructed from the human-verified valid action set $\mathcal{A}_t^*$, 
We denote this teacher-side guidance as: 
\begin{equation}
g_t = h(\mathcal{A}_t^*, \hat{a}_t),
\end{equation}
where $h(\cdot)$ specifies how the annotated valid actions are converted into the teacher-side action guidance. In this way, the teacher distribution is conditioned on $\pi_T(\cdot \mid x_t, g_t)$, whereas the student policy remains $\pi_\theta(\cdot \mid x_t)$.  Note that the guidance $g_t$ is derived only from the current-step human-verified valid action set $\mathcal{A}_t^*$ and is used exclusively on the teacher side during training. It is never provided to the student model and is not used at inference time. Therefore, this framework can be viewed as a form of training-time privileged guidance, following the learning-using-privileged-information paradigm~\citep{vapnik2009new,vapnik2015learning}, and as a privileged-information extension of on-policy distillation~\citep{lopez2015unifying,agarwal2024onpolicy}.
Different guidance variants $h(\cdot)$ correspond to different ways of mapping $\mathcal{A}_t^*$ into $g_t$. We consider three variants: Single-GT, Multi-GT and Most-Matched-GT Guided OPD in this paper.

\textbf{(1) Single-GT Guided OPD}. One valid action is randomly sampled from $\mathcal{A}_t^*$ and provided to the teacher as the reference action:
\begin{equation}
g_t = a_t^{*(k)}, 
\quad 
k \sim \mathrm{Uniform}\{1,\dots,K\}.
\end{equation}
This gives the teacher additional task constraints, but the randomly selected ground truth may be far from the student's current output, leading to a mismatch between the selected reference action and the student’s current output.

\textbf{(2) Multi-GT Guided OPD}. The full valid action set is provided to the teacher:
\begin{equation}
g_t = \mathcal{A}_t^*.
\end{equation}
This exposes the teacher to all annotated valid actions for the current state. However, it may also increase the complexity of the teacher context and lead to a more diffuse likelihood estimate across multiple possible actions.

\textbf{(3) Most-Matched-GT Guided OPD}. 
Given the student-generated output $\hat{y}_t$, we select the valid action that is most matched to the student's current on-policy behavior:
\begin{equation}
g_t = a_t^\dagger
=
\arg\max_{a^* \in \mathcal{A}_t^*}
\phi_{\mathrm{gui}}(\hat{y}_t, a^*),
\label{eq:most_matched_gt}
\end{equation}
where $\phi_{\mathrm{gui}}$ is the unified GUI action matching function defined as: 

\begin{equation}
\phi_{\mathrm{gui}}(\hat{y}_t, a^*) =
\frac{
w_{\tau} m_{\tau}
+
\mathbb{I}[b^*\neq \emptyset] w_p m_p
+
\mathbb{I}[v^*\neq \emptyset] w_v m_v
}{
w_{\tau}
+
\mathbb{I}[b^*\neq \emptyset] w_p
+
\mathbb{I}[v^*\neq \emptyset] w_v
}.
\label{eq:gui_action_matcher}
\end{equation}
$m_{\tau}=\mathbb{I}[\hat{\tau}_t=\tau^*]$ measures action-type correctness, $m_p\in[0,1]$ measures the alignment between the predicted position $\hat{p}_t$ and the target bounding box $b^*$, and $m_v\in[0,1]$ measures exact or partial matching between the predicted value $\hat{v}_t$ and the annotated value $v^*$. The indicator terms ensure that position and value scores are only considered when the corresponding annotations are available. Details of $m_{\tau}$, $m_p$, $m_v$, and corresponding coefficients $w_{\tau}$, $w_p$, $w_v$ are provided in Appendix~\ref{app:reward_details}. If the student output cannot be parsed into a valid GUI action, all candidate actions receive zero matching scores, and we fall back to a deterministic reference action from $\mathcal{A}_t^*$. This selection mechanism provides the teacher with a reference action that is aligned with the student's current on-policy output. Compared with Single-GT, it reduces the mismatch caused by random or fixed reference selection. Compared with Multi-GT, it provides a more focused reference signal while avoiding an overly complex teacher context.

\begin{table}[!t]
\caption{Controlled ablations with Qwen3-VL-2B under the common Lite-Dataset protocol (five-run mean). ScreenSpot-Pro values here use Lite-Dataset rather than the ShowUI-Desktop-8K protocol in Table~\ref{tab:screenspot_pro}.} 
\label{tab:mul_sub}
\centering
\scriptsize
\setlength{\tabcolsep}{2.5pt}
\resizebox{\columnwidth}{!}{%
\begin{tabular}{lccccc}
\hline
\multirow{2}{*}{Method} & \multirow{2}{*}{Multi-Sol.} & \multirow{2}{*}{Dual-Lv.} & \multicolumn{3}{c}{Benchmark} \\ \cline{4-6} 
  &  &  & SS-Pro & OSWorld & Lite-Bench \\ \hline
 Qwen3-VL-2B \citep{Qwen3-VL} & - & - & 40.16 & 6.04 & 32.35 \\
 + SFT & - & - & 31.56 & 3.25 & 37.96 \\
 + SFT + GRPO & no & no & 33.47 & 3.77 & 42.65 \\ \hline
 + OPD & - & - & 41.87 & 9.32 & 51.47 \\
 + Single-Guided OPD & - & - & 41.68 & 11.04 & 51.76 \\
 + Multi-Guided OPD & - & - & 42.25 & 11.11 & 52.94 \\
 + Most-Matched-Guided OPD & - & - & 42.50 & 11.56 & 55.88 \\ \hline
 \multirow{4}{*}{\begin{tabular}[c]{@{}l@{}}+ Most-Matched\\\quad Guided OPD + GRPO\end{tabular}} & no & no & 42.37 & 6.56 & 52.94 \\
  & no & yes & 42.50 & 8.13 & 59.20 \\
  & yes & no & 42.69 & 11.84 & 60.58 \\
  & yes & yes & \textbf{43.13} & \textbf{13.24} & \textbf{63.13} \\ \hline
\end{tabular}%
}
\vspace{-6pt}
\end{table}

\textbf{Full-vocabulary reverse-KL OPD}. After obtaining $g_t$, we construct the input of teacher model as
\begin{equation}
\tilde{x}_t =
\begin{aligned}
[&q, s_{t-w}, y_{t-w}, \dots, s_{t-1}, y_{t-1}, s_t,\\
&\texttt{reference\_action}=g_t].
\end{aligned}
\end{equation}
The student samples a completion $\hat{y}_t\sim\pi_\theta(\cdot\mid x_t)$. The sampled token sequence and the discrete teacher-side guidance $g_t$ are treated as fixed during the optimization step. Let $\mathcal{M}_t$ denote the non-masked completion positions. At each valid position $j\in\mathcal{M}_t$, we compute the full-vocabulary student and teacher distributions:
\begin{equation}
p_{\theta,j}=\pi_\theta(\cdot\mid x_t,\hat{y}_{t,<j}),
\qquad
p_{T,j}^{g_t}=\pi_T(\cdot\mid \tilde{x}_t,\hat{y}_{t,<j}).
\end{equation}
Following GKD~\citep{agarwal2024onpolicy}, the objective is the full-vocabulary, per-prefix divergence:
\begin{equation}
\mathcal{L}_{\mathrm{OPD}}
=
\frac{1}{|\mathcal{M}_t|}
\sum_{j\in\mathcal{M}_t}
D_{\beta}\left(
p_{\theta,j}\,\middle\|\,\operatorname{sg}\left[p_{T,j}^{g_t}\right]
\right),
\qquad \beta=1,
\end{equation}
where $\operatorname{sg}[\cdot]$ denotes stop-gradient (identity in the forward pass, with zero gradient), and $D_{\beta}$ denotes the generalized JSD/KL family used by our implementation. Under our parameterization, $\beta=1$ selects reverse KL:
\begin{equation}
D_{1}(p\|q)=D_{\mathrm{KL}}(p\|q) 
=\sum_{v}p(v)\log\frac{p(v)}{q(v)}.
\end{equation}
The loss is computed over the full vocabulary, rather than only the sampled token, with gradients flowing only through the student distribution. See Appendix~\ref{app:gkd_training} for implementation details.

\begin{figure*}[t]
  \centering
  \includegraphics[width=0.78\textwidth]{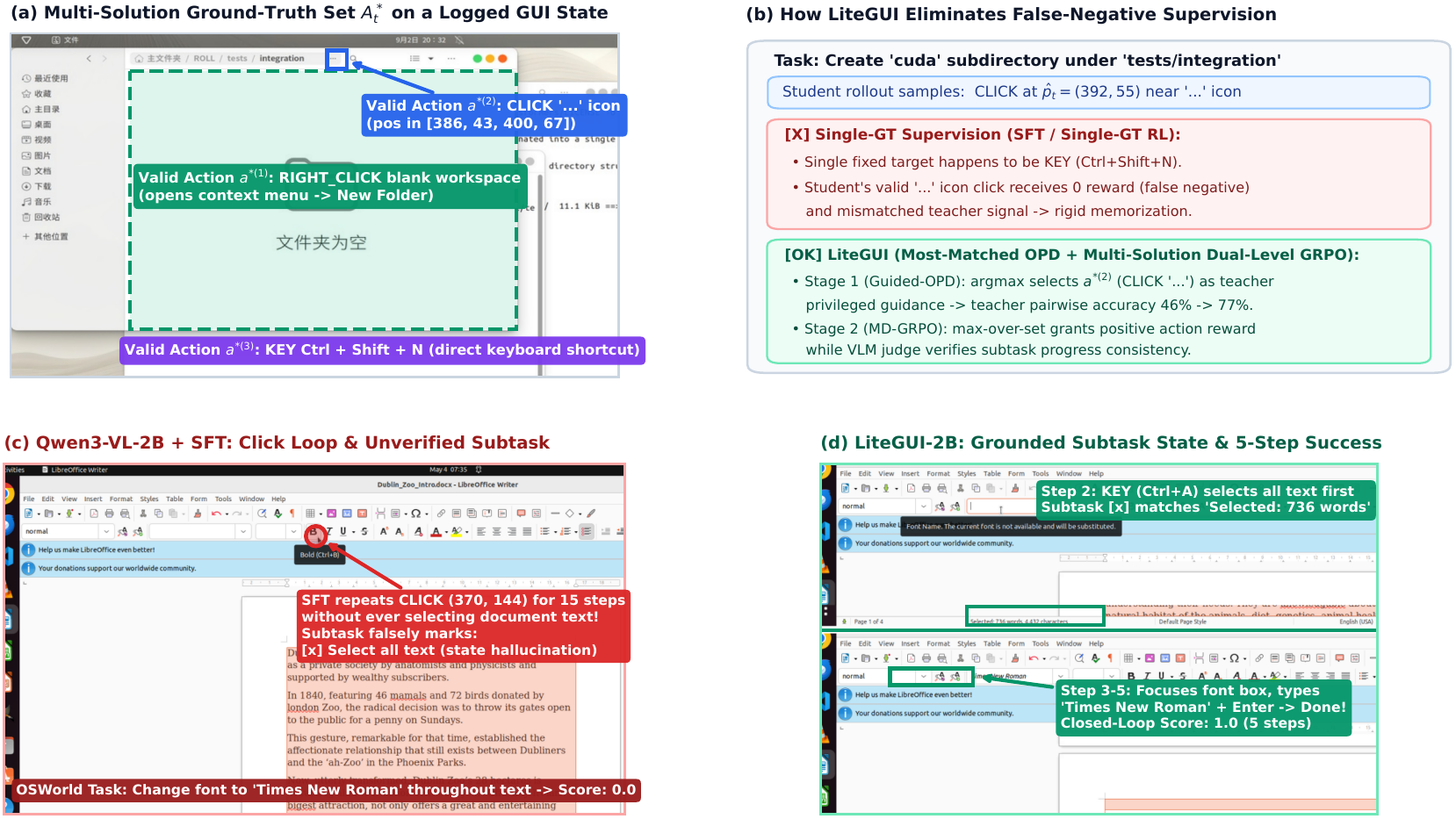}
  \vspace{-4pt}
  \caption{Qualitative comparison on logged multi-solution supervision and closed-loop OSWorld execution. \textbf{(a--b)} On an Ubuntu file-manager state, three distinct actions validly create a new folder (context menu, toolbar icon, or hotkey); Most-Matched Guided-OPD and MD-GRPO align supervision with the student's sampled valid mode instead of penalizing non-template paths. \textbf{(c)} On an OSWorld LibreOffice Writer task (\emph{``Change the font to `Times New Roman' throughout the text''}), Qwen3-VL-2B + SFT hallucinates subtask completion (\texttt{[x] Select all text}) without selecting the document body and repeats the same toolbar click for 15 steps (Score: $0.0$). \textbf{(d)} LiteGUI-2B selects all 736 words via \texttt{Ctrl+A} (verified against the bottom status bar), updates its subtask state coherently, and applies \texttt{Times New Roman} in 5 steps (Score: $1.0$).}
  \label{fig:qualitative_main}
  \vspace{-6pt}
\end{figure*}

\subsection{Multi-solution Dual-level GRPO}
\label{sec:MD_GRPO}

We further optimize the student using GRPO. Unlike standard single-answer tasks, GUI control exhibits strong multi-solution equivalence: under the same GUI state, multiple actions may validly advance the task. For example, a file can be opened by double-clicking it or by selecting it and pressing Enter; a web page can be reached by clicking a shortcut or directly typing the URL. Therefore, a reward based on a single ground-truth action may incorrectly penalize valid alternative actions.

We focus on two reward-design questions that are central to GUI agent training: how to avoid false negatives caused by single-reference supervision, and how to evaluate whether the model understands the current task stage and can re-plan when necessary. We address these questions using the \textbf{multi-solution action reward} and \textbf{planning reward}. For each logged input $x_t$, the behavior policy samples $G$ candidate outputs:
\begin{equation}
\{\hat{y}_t^{(1)}, \hat{y}_t^{(2)}, \dots, \hat{y}_t^{(G)}\}
\sim
\pi_{\theta_{\mathrm{old}}}(\cdot \mid x_t).
\end{equation}
Training uses logged GUI states rather than live environment rollouts: all
candidates are scored at the same state using the trajectory context and
human-verified annotations.
The total reward for the $i$-th candidate is
\begin{equation}
\begin{aligned}
R_t^{(i)} ={}&
\lambda_{ms} R_{\mathrm{ms}}(\hat{y}_t^{(i)}, \mathcal{A}_t^*)
+\lambda_{sub} R_{\mathrm{sub}}(\hat{y}_t^{(i)})\\
&+\lambda_{format/length} R_{\mathrm{format/length}}(\hat{y}_t^{(i)}),
\end{aligned}
\end{equation}
where $R_{\mathrm{ms}}$ is the multi-solution GUI action reward,
$R_{\mathrm{sub}}$ is the planning reward, and
$R_{\mathrm{format/length}}$ is the JSON-aware soft overlength reward. $\lambda_{ms}$, $\lambda_{sub}$ and $\lambda_{format/length}$ are the coefficients for each reward term. Detailed hyperparameter settings can be found in Appendix~\ref{app:grpo_training}. 

\textbf{Multi-solution Action Reward}. 
Given the multi-solution action set $\mathcal{A}_t^*$, we compare the model output against all valid candidate actions and take the maximum score as:
\begin{equation}
R_{\mathrm{ms}}(\hat{y}_t, \mathcal{A}_t^*)
=
\max_{a^* \in \mathcal{A}_t^*}
\phi_{\mathrm{gui}}(\hat{y}_t, a^*).
\label{eq:multi_solution_reward}
\end{equation}

where $\phi_{\mathrm{gui}}$ is the unified GUI action matching function defined in Eq.~\ref{eq:gui_action_matcher}. This changes the reward criterion from ``does the model reproduce this specific annotated action?'' to ``does the model match any acceptable action?'' Since a single annotation is usually only one sample from the space of valid GUI actions, max-over-solutions substantially reduces false negative rewards caused by fixed annotation paths. From an optimization perspective, single-ground-truth rewards can incorrectly convert differences among equivalent action paths into negative feedback, discouraging exploration of valid alternatives. In contrast, the multi-solution reward gives positive feedback when the model output matches any annotated acceptable action for the current logged state.

\textbf{History-conditioned Planning-Quality Reward}. 
The multi-solution action reward provides action-level supervision by matching each candidate output against annotated acceptable actions for the current logged GUI state. For multi-step GUI tasks, a complementary planning signal can assess whether a proposed subtask plan reflects task progress and the intended next step. We therefore introduce a history-conditioned, per-state planning-quality reward for this purpose. The reward scores the current candidate subtask plan $c_t$, which compactly represents task progress, using the available trajectory context. It therefore evaluates intermediate plan quality at the current logged state rather than the outcome of executing a candidate action. 
The frozen VLM judge uses the task, available screenshot history, and previous model outputs as context and returns a discrete rubric score:
\begin{equation}
\begin{aligned}
R_{\mathrm{sub}} ={}& f_{\mathrm{judge}}(q, s_{t-w:t}, y_{t-w:t-1}, c_t),\\
&R_{\mathrm{sub}} \in \{0.0, 0.3, 0.5, 0.8, 1.0\}.
\end{aligned}
\end{equation}
It evaluates the plan along four dimensions: task relevance and adaptability, visual grounding, decomposition granularity, and state consistency. In particular, when the latest screenshot indicates that the agent is stuck, in an error state, or in a loop, the judge is instructed to assign a low score unless the subtask plan explicitly reflects the abnormal state and proposes a corrective next step. Together, the two rewards provide action-level supervision and history-conditioned plan-quality supervision. Candidate actions are not executed for reward computation, and the reward uses neither sampled future states nor terminal task-success feedback. It is therefore a local surrogate for plan quality, not an estimate of long-term return, verified error recovery, or temporal credit assignment.

\textbf{GRPO Objective.} For the $G$ candidate outputs sampled from the same input, we normalize rewards within the group to obtain the relative advantage:
\begin{equation}
\hat{A}^{(i)}_t=
\frac{
R^{(i)}_t-\mathrm{mean}(\{R^{(j)}_t\}_{j=1}^{G})
}{
\mathrm{std}(\{R^{(j)}_t\}_{j=1}^{G})+\epsilon_{\mathrm{adv}}
}.
\end{equation}
Here $\epsilon_{\mathrm{adv}}$ is the numerical stabilizer for reward
normalization and is distinct from the clipping bounds
$\epsilon_{\mathrm{low}}$ and $\epsilon_{\mathrm{high}}$.
We then optimize the policy using the GRPO clipped objective:
\begin{equation}
\begin{aligned}
\mathcal{L}_{\mathrm{GRPO}} ={}&
-\frac{1}{G}\sum_{i=1}^{G}
\min\Big(\rho^{(i)}_t\hat{A}^{(i)}_t,\\
&\qquad\mathrm{clip}\big(\rho^{(i)}_t,1-\epsilon_{\mathrm{low}},1+\epsilon_{\mathrm{high}}\big)
\hat{A}^{(i)}_t\Big)\\
&+\beta_{\mathrm{KL}}D_{\mathrm{KL}}(\pi_\theta\|\pi_{\mathrm{ref}}),
\end{aligned}
\end{equation}
where the sequence-level policy ratio is computed from the mean token
log-probability difference:
\begin{equation}
\begin{aligned}
\rho^{(i)}_t={}&
\exp\!\Bigg[
\frac{1}{|\mathcal{M}^{(i)}_t|}
\sum_{j\in\mathcal{M}^{(i)}_t}
\Big(
\log \pi_\theta(\hat{y}^{(i)}_{t,j}\mid x_t,\hat{y}^{(i)}_{t,<j})\\
&-\log \pi_{\theta_{\mathrm{old}}}(\hat{y}^{(i)}_{t,j}\mid x_t,\hat{y}^{(i)}_{t,<j})
\Big)
\Bigg].
\end{aligned}
\end{equation}
Here, $\mathcal{M}^{(i)}_t$ denotes the valid completion-token positions,
$\pi_{\theta_{\mathrm{old}}}$ is the behavior policy used to sample
candidates, and $\pi_{\mathrm{ref}}$ is the reference model. Detailed hyperparameter settings are in Appendix~\ref{app:grpo_training}.

\begin{table*}[!t]
\caption{Performance comparison on ScreenSpot-Pro. LiteGUI models here follow the ShowUI-Desktop-8K training protocol.}
\scriptsize
\label{tab:screenspot_pro}
\resizebox{\textwidth}{!}{
\begin{tabular}{lccccccccccccc}
\hline
 & \multicolumn{13}{c}{ScreenSpot-Pro Accuracy (\%)} \\ \cline{2-14} 
 & \multicolumn{2}{c}{CAD} & \multicolumn{2}{c}{Dev} & \multicolumn{2}{c}{Creative} & \multicolumn{2}{c}{Scientific} & \multicolumn{2}{c}{Office} & \multicolumn{2}{c}{OS} &  \\
\multirow{-3}{*}{Model} & Text & Icon & Text & Icon & Text & Icon & Text & Icon & Text & Icon & Text & Icon & \multirow{-2}{*}{Avg.} \\ \hline
\multicolumn{14}{c}{Proprietary Models} \\ \hline
GPT-4o \citep{hurst2024gpt}& 2 & 0 & 1.3 & 0 & 1 & 0 & 2.1 & 0 & 1.1 & 0 & 0 & 0 & 0.8 \\
Claude Computer Use \citep{claude} & 14.5 & 3.7 & 22 & 3.9 & 25.9 & 3.4 & 33.9 & 15.8 & 30.1 & 16.3 & 11 & 4.5 & 17.1 \\ \hline
\multicolumn{14}{c}{{\color[HTML]{333333} General Open-source Models}} \\ \hline
Qwen2.5-VL-3B \citep{Qwen2.5-VL}& 9.1 & 7.3 & 22.1 & 1.4 & 26.8 & 2.1 & 38.2 & 7.3 & 33.9 & 15.1 & 10.3 & 1.1 & 16.1 \\
Qwen2.5-VL-7B \citep{Qwen2.5-VL}& 16.8 & 1.6 & 46.8 & 4.1 & 35.9 & 7.7 & 49.3 & 7.3 & 52.5 & 20.8 & 37.4 & 6.7 & 26.8 \\
Qwen3-VL-2B \citep{Qwen3-VL} & 52.79 & 15.62 & 50.0 & 10.34 & 48.48 & 18.18 & 59.02 & 23.63 & 58.19 & 33.96 & 51.40 & 21.34 & 40.16 \\
Qwen3-VL-32B \citep{Qwen3-VL} & 71.06 & 35.93 & 76.62 & 24.13 & 72.72 & 29.37 & 84.72 & 31.81 & 85.31 & 54.71 & 68.22 & 15.73 & 58.57 \\ \hline
\multicolumn{14}{c}{GUI-specific Models (SFT)} \\ \hline
SeeClick-9.6B \citep{cheng2024seeclick}& 2.5 & 0 & 0.6 & 0 & 1 & 0 & 3.5 & 0 & 1.1 & 0 & 2.8 & 0 & 1.1 \\
CogAgent-18B \citep{hong2024cogagent}& 7.1 & 3.1 & 14.9 & 0.7 & 9.6 & 0 & 22.2 & 1.8 & 13 & 0 & 5.6 & 0 & 7.7 \\
Aria-UI \citep{ariaui}& 7.6 & 1.6 & 16.2 & 0 & 23.7 & 2.1 & 27.1 & 6.4 & 20.3 & 1.9 & 4.7 & 0 & 11.3 \\
OS-Atlas-7B \citep{wu2024atlas}& 12.2 & 4.7 & 33.1 & 1.4 & 28.8 & 2.8 & 37.5 & 7.3 & 33.9 & 5.7 & 27.1 & 4.5 & 18.9 \\
ShowUI-2B \citep{lin2024showui}& 2.5 & 0 & 16.9 & 1.4 & 9.1 & 0 & 13.2 & 7.3 & 15.3 & 7.5 & 10.3 & 2.2 & 7.7 \\
UGround-V1-7B\citep{gou2025uground} & 15.8 & 1.2 & 51.9 & 2.8 & 47.5 & 9.7 & 57.6 & 14.5 & 60.5 & 13.2 & 38.3 & 7.9 & 31.1 \\
UI-TARS-2B \citep{qin2025ui}& 17.8 & 4.7 & 47.4 & 4.1 & 42.9 & 6.3 & 56.9 & 17.3 & 50.3 & 17 & 21.5 & 5.6 & 27.7 \\
UI-TARS-72B \citep{qin2025ui}& 18.8 & 12.5 & 62.9 & 17.2 & 57.1 & 15.4 & 64.6 & 20.9 & 63.3 & 26.4 & 42.1 & 15.7 & 38.1 \\
JEDI-3B \citep{xie2025scalingcomputerusegroundinguser}& 27.4 & 9.4 & 61 & 13.8 & 53.5 & 8.4 & 54.2 & 18.2 & 64.4 & 32.1 & 38.3 & 9 & 36.1 \\
JEDI-7B \citep{xie2025scalingcomputerusegroundinguser}& 38 & 14.1 & 42.9 & 11 & 50 & 11.9 & 72.9 & 25.5 & 75.1 & 47.2 & 33.6 & 16.9 & 39.5 \\
GUI-Actor-7B \citep{wu2025gui}& - & - & - & - & - & - & - & - & - & - & - & - & 44.6 \\ \hline
\multicolumn{14}{c}{GUI-specific Models (RL)} \\ \hline
UI-R1-E-3B \citep{lu2025ui}& 37.1 & 12.5 & 46.1 & 6.9 & 41.9 & 4.2 & 56.9 & 21.8 & 65 & 26.4 & 32.7 & 10.1 & 33.5 \\
GUI-R1-3B \citep{luo2025gui} & 26.4 & 7.8 & 33.8 & 4.8 & 40.9 & 5.6 & 61.8 & 17.3 & 53.6 & 17 & 28.1 & 5.6 & - \\
GUI-R1-7B \citep{luo2025gui}& 23.9 & 6.3 & 49.4 & 4.8 & 38.9 & 8.4 & 55.6 & 11.8 & 58.7 & 26.4 & 42.1 & 16.9 & - \\
InfigUI-R1-3B \citep{liu2025infigui} & 33 & 14.1 & 51.3 & 12.4 & 44.9 & 7 & 58.3 & 20 & 65.5 & 28.3 & 43.9 & 12.4 & 35.7 \\
GUI-G1-3B \citep{zhou2025guig1}& 39.6 & 9.4 & 50.7 & 10.3 & 36.6 & 11.9 & 61.8 & 30 & 67.2 & 32.1 & 23.5 & 10.6 & 37.1 \\
SE-GUI-3B \citep{yuan2025enhancing}& 38.1 & 12.5 & 55.8 & 7.6 & 47 & 4.9 & 61.8 & 16.4 & 59.9 & 24.5 & 40.2 & 12.4 & 35.9 \\
\hline
\multicolumn{14}{c}{Ours} \\ \hline
LiteGUI-2B & 61.92 & 20.31 & 59.09 & 14.48 & 59.59 & 22.37 & 68.75 & 25.45 & 65.53 & 28.30 & 59.81 & 24.71 & 46.86 \\
LiteGUI-30B-A3B & 74.61 & 20.31 & 76.62 &29.65 & 70.20 & 21.67 & 79.16 & 33.63 & 80.79 & 50.94 & 79.43 & 39.32 & \textbf {58.95} \\ \hline
\end{tabular}
}
\vspace{-8pt}
\end{table*}

\section{Experiments}
\label{sec:experiments}
\subsection{Experimental Setup}
We implement LiteGUI using two representative backbones: Qwen3-VL-2B and Qwen3-VL-30B-A3B, and employ Qwen3-VL-32B as the teacher model.
We construct \textbf{Lite-Dataset} via our pipeline (Figure~\ref{fig:acc}): 30K trajectories segmented into $\sim$70K three-step windowed samples with 11K verified multi-solution annotations (Appendices~\ref{app:data_pipeline}--\ref{app:datasets}). Two protocols are distinguished: (1) \textbf{Controlled ablations} (Table~\ref{tab:mul_sub}) train Qwen3-VL-2B on Lite-Dataset (3K multi-solution subset for Guided-OPD) and evaluate the same checkpoint on all three benchmarks; (2) \textbf{Main comparisons} train OSWorld (Table~\ref{tab:osworld}) and Lite-Bench (Table~\ref{tab:lite_benchmark}) models on Lite-Dataset, while ScreenSpot-Pro (Table~\ref{tab:screenspot_pro}) augments ShowUI-Desktop-8K~\citep{lin2024showui} with multi-solution annotations (hence 43.13\% in Table~\ref{tab:mul_sub} vs.\ 46.86\% in Table~\ref{tab:screenspot_pro}). MD-GRPO uses frozen Qwen3-VL-32B-Instruct as the training-time planning judge (90.0\% human agreement, $\kappa{=}0.79$; Appendix~\ref{app:judge_prompt}) on A100 GPUs (Appendices~\ref{app:gkd_training}--\ref{app:grpo_training}).

All evaluations are \textbf{vision-only} (screenshots only, no accessibility trees or DOM) across OSWorld~\citep{OSWorld}, ScreenSpot-Pro~\citep{li2025screenspotpro}, and Lite-Bench (160 tasks up to 18 steps: 75 web, 38 terminal, 47 GUI-only file tasks). In Tables~\ref{tab:mul_sub}--\ref{tab:lite_benchmark}, all our models and controlled baselines use the exact same standalone agent wrapper (three-step screenshot context) without external runtime loop-breaker prompts, auxiliary verifiers, or test-time search. Lite-Bench outcomes are audited by GPT-5.4 (93.9\% human agreement; Appendix~\ref{app:litebench}). All our results are five-run means.

\subsection{Effectiveness of LiteGUI}

We first compare LiteGUI with SFT, GRPO, and OPD, as well as ablations of individual components, across all three benchmarks in Table~\ref{tab:mul_sub}. All methods use the same Qwen3-VL-2B backbone and Lite-Dataset, and the reported results are averaged over five independent runs. Under this controlled protocol, the full LiteGUI configuration obtains the highest values among the listed variants on all three benchmarks. The lower SFT scores are consistent with possible overfitting in this setting, although the present experiments do not isolate the individual causes.

A closer examination of the ablations shows that Guided-OPD outperforms vanilla OPD, while Most-Matched Guided-OPD achieves the best, demonstrating the benefit of using the guided supervision. Moreover, incorporating either multi-solution or dual-level rewards in GRPO further improves performance, while combining all components achieves the strongest overall results. Notably, OSWorld and Lite-Bench contain more multi-step GUI tasks, whereas ScreenSpot-Pro focuses primarily on visual grounding and single-step interaction. Consequently, LiteGUI yields larger gains on OSWorld and Lite-Bench in this evaluation, which is consistent with the usefulness of logged-state plan-quality and multi-solution action signals for multi-step execution.

Across 50 multi-solution validation states (8 rollouts/state; Appendix~\ref{app:guided_opd_analysis}), privileged guidance raises teacher pairwise preference accuracy from 46\% to 77\%, boosting teacher scores on both selected (0.596$\to$0.980) and \emph{unselected} valid actions (0.314$\to$0.532). On the student side, Most-Matched Guided-OPD raises \texttt{ValidRate@8} (0.518$\to$0.573) while \emph{increasing} distinct valid modes (\texttt{UniqueValid@8}: 0.52$\to$0.66) and \texttt{Coverage@8} (0.26$\to$0.33); MD-GRPO further lifts \texttt{ValidRate@8} to 0.638 while preserving both diversity metrics, ruling out mode collapse. Figure~\ref{fig:qualitative_main} visually illustrates how multi-solution matching preserves equivalent desktop interaction modes and prevents repetitive click loops (further examples in Appendix~\ref{app:qualitative}).

\subsection{Comparison with Representative Published Methods}
\begin{table}[!t]
\centering
\caption{Performance comparison on OSWorld (screenshot-only setting for LiteGUI; ``Params (Act.)'' denotes total / active parameters).}
\scriptsize
\setlength{\tabcolsep}{1.8pt}
\label{tab:osworld}
\begin{tabular}{lcccc}
\hline
Model & Params (Act.) & Type & Open & \begin{tabular}[c]{@{}c@{}}Success Rate\\ (Avg \%)\end{tabular} \\ \hline
Claude-sonnet-4-5 \citep{claude}& -- & General & False & 42.9 \\
Doubao-1.5-pro \citep{guo2025seed15vltechnicalreport}& -- & Specialized & False & 39.0 \\
GPT-o3 \citep{gpt_o3}& -- & General & False & 9.1 \\ \hline
Qwen2.5-VL-32B-Instruct \citep{Qwen2.5-VL}& 32B (32B) & General & True & 3.0 \\
Qwen2.5-VL-72B-Instruct \citep{Qwen2.5-VL}& 72B (72B) & General & True & 4.4 \\
OpenCUA-Qwen2-7B \citep{wang2025opencuaopenfoundationscomputeruse}& 7B (7B) & Specialized & True & 19.9 \\
OpenCUA-7B \citep{wang2025opencuaopenfoundationscomputeruse}& 7B (7B) & Specialized & True & 24.3 \\
UI-TARS-1.5-7B \citep{qin2025ui} & 7B (7B) & Specialized & True & 24.5 \\
UI-TARS-72B-DPO \citep{qin2025ui}& 72B (72B) & Specialized & True & 24.0 \\ \hline
Qwen3-VL-2B-Instruct \citep{Qwen3-VL}& 2B (2B) & General & True & 6.04 \\
Kimi-VL-A3B \citep{kimiteam2025kimivltechnicalreport}& 16B (2.8B) & General & True & 9.7 \\
Qwen3-VL-30B-A3B-Instruct \citep{Qwen3-VL} & 30B (3B) & General & True & 16.7 \\
OpenCUA-A3B \citep{wang2025opencuaopenfoundationscomputeruse}& 16B (2.8B) & Specialized & True & 16.9 \\
\textbf{LiteGUI-2B (Ours)} & \textbf{2B (2B)} & Specialized & True & \textbf{13.24} \\
\textbf{LiteGUI-30B-A3B (Ours)} & \textbf{30B (3B)} & Specialized & True & \textbf{22.7} \\ \hline
\end{tabular}
\vspace{-6pt}
\end{table}

\begin{table}[!t]
\caption{Performance comparison on Lite-Bench across bare backbones, post-training baselines, and LiteGUI.}
\label{tab:lite_benchmark}
\scriptsize
\centering
\setlength{\tabcolsep}{3.5pt}
\begin{tabular}{lcccc}
\hline
\multirow{2}{*}{Model / Post-training Method} & \multicolumn{4}{c}{Success Rate (\%)} \\ \cline{2-5} 
 & File System & Web & Terminal & All\\ \hline
Qwen3-VL-2B \citep{Qwen3-VL} & 23.00 & 35.00 & 41.00 & 32.35\\
\quad + SFT & -- & -- & -- & 37.96\\
\quad + SFT + GRPO & -- & -- & -- & 42.65\\
\quad + OPD & -- & -- & -- & 51.47\\
\quad + Most-Matched-Guided OPD & -- & -- & -- & 55.88\\
\textbf{LiteGUI-2B (Ours)} & \textbf{40.42} & \textbf{69.33} & \textbf{78.94} & \textbf{63.13}\\ \hline
Qwen3-VL-30B-A3B \citep{Qwen3-VL} & 50.00 & 53.00 & 92.00 & 61.34\\
Qwen3-VL-32B (Teacher) \citep{Qwen3-VL} & 85.00 & 82.00 & 95.00 & 85.88 \\
\textbf{LiteGUI-30B-A3B (Ours)} & \textbf{89.36} & \textbf{84.00} & \textbf{97.36} & \textbf{88.75}\\ \hline
\end{tabular}
\vspace{-8pt}
\end{table}

Tables~\ref{tab:screenspot_pro}--\ref{tab:lite_benchmark} compare LiteGUI with published methods under the unified benchmark protocols defined in Sec.~4.1 (all evaluated under the same standalone vision-only wrapper without runtime loop-breaker prompts). On ScreenSpot-Pro (Table~\ref{tab:screenspot_pro}), LiteGUI-2B reaches 46.86\% average accuracy and LiteGUI-30B-A3B reaches 58.95\%, surpassing much larger dense models such as UI-TARS-72B (38.1\%).

On OSWorld (Table~\ref{tab:osworld}), LiteGUI-2B reaches 13.24\% (more than doubling its 6.04\% backbone), while LiteGUI-30B-A3B reaches 22.7\%---leading all 2B--3B active-parameter models (vs.\ OpenCUA-A3B at 16.9\%) and approaching dense 7B--72B systems (UI-TARS-1.5-7B at 24.5\%, UI-TARS-72B-DPO at 24.0\%) at a fraction of their active compute. A per-trajectory audit on a fully logged 361-task 30B-A3B run (22.10\% overall; $+5.71$\,pp paired gain over the backbone on 359 matched tasks, $p{=}0.001$; Figure~\ref{fig:qualitative_main} and Appendix~\ref{app:osworld_trajectory_audit}) shows 100\% valid JSON across 4,603 turns and full-score completions doubling in 13--15 step rollouts ($11\!\to\!22$), while residual failures stem mainly from 15-step horizon exhaustion (45.6\%) and local action loops (35.2\%). On Lite-Bench (Table~\ref{tab:lite_benchmark}), LiteGUI-2B reaches 63.13\% (vs.\ 37.96\% SFT, 42.65\% SFT+GRPO, and 55.88\% Guided-OPD), and LiteGUI-30B-A3B reaches 88.75\%, surpassing the 32B teacher (85.88\%).

\textbf{Deployment Efficiency.} With identical three-step screenshot contexts (Appendix~\ref{app:efficiency}), LiteGUI-2B runs at 103.0 tok/s (4.5\,GB VRAM) and LiteGUI-30B-A3B at 91.5 tok/s (18\,GB) on one A100, whereas the Qwen3-VL-32B teacher needs $2{\times}$A100 at 27.3 tok/s. On one RTX 4090, LiteGUI-2B/30B-A3B reach 81.2/17.7 tok/s while 32B (AWQ) runs out of memory. Higher offline training cost (17 vs.\ 4 GPU-hours for Guided-OPD vs.\ SFT; 1.2K GPU-hours and 120 human hours overall) is strictly one-time, with zero teacher/judge overhead at inference. These results highlight the central design principle of LiteGUI: shifting computational complexity from repeated inference to one-time offline post-training. Although our training pipeline incurs additional computational and annotation costs, the resulting agents operate independently at deployment, without requiring teacher models, reward judges, or test-time search.

\paragraph{Limitations.}
Because LiteGUI optimizes logged GUI states rather than live environment rollouts, it does not directly optimize long-horizon returns beyond the three-step window or verify terminal states in unexplored menus; moreover, $\mathcal{A}_t^*$ is a finite approximation of valid actions and the frozen VLM judge adds offline training cost (see Appendix~\ref{app:limitations}).

\section{Conclusion}
We presented LiteGUI, a post-training framework for lightweight vision-only GUI agents combining Most-Matched Guided-OPD with Multi-Solution Dual-Level GRPO over logged GUI states, alongside Lite-Dataset and Lite-Bench. Across ScreenSpot-Pro, OSWorld, and Lite-Bench, LiteGUI preserves multi-path action diversity and transfers offline supervision to closed-loop desktop execution with zero inference-time judge overhead.

{\small
\bibliographystyle{ieeenat_fullname}
\bibliography{ref}

\begin{thebibliography}{45}
\providecommand{\natexlab}[1]{#1}
\providecommand{\url}[1]{\texttt{#1}}
\expandafter\ifx\csname urlstyle\endcsname\relax
  \providecommand{\doi}[1]{doi: #1}\else
  \providecommand{\doi}{doi: \begingroup \urlstyle{rm}\Url}\fi

\bibitem[Agarwal et~al.(2024)Agarwal, Vieillard, Zhou, Stanczyk, Ramos~Garea,
  Geist, and Bachem]{agarwal2024onpolicy}
Rishabh Agarwal, Nino Vieillard, Yongchao Zhou, Piotr Stanczyk, Sabela
  Ramos~Garea, Matthieu Geist, and Olivier Bachem.
\newblock On-policy distillation of language models: Learning from
  self-generated mistakes.
\newblock In \emph{International Conference on Learning Representations}, 2024.

\bibitem[Agashe et~al.(2025)Agashe, Wong, Tu, Yang, Li, and
  Wang]{agashe2025agent}
Saaket Agashe, Kyle Wong, Vincent Tu, Jiachen Yang, Ang Li, and Xin~Eric Wang.
\newblock Agent s2: A compositional generalist-specialist framework for
  computer use agents, 2025.
\newblock \emph{URL https://arxiv. org/abs/2504.00906}, 2:\penalty0 10--16,
  2025.

\bibitem[Alansari and Luqman(2026)]{alansari2026large}
Aisha Alansari and Hamzah Luqman.
\newblock Large language models hallucination: A comprehensive survey.
\newblock \emph{Computer Science Review}, 61:\penalty0 100970, 2026.

\bibitem[anthropic(2026)]{claude}
anthropic.
\newblock claude-3-family.
\newblock \url{https://www.anthropic.com/news/claude-3-family}, 2026.

\bibitem[Bai et~al.(2023)Bai, Bai, Yang, Wang, Tan, Wang, Lin, Zhou, and
  Zhou]{Qwen-VL}
Jinze Bai, Shuai Bai, Shusheng Yang, Shijie Wang, Sinan Tan, Peng Wang, Junyang
  Lin, Chang Zhou, and Jingren Zhou.
\newblock Qwen-vl: A versatile vision-language model for understanding,
  localization, text reading, and beyond.
\newblock \emph{arXiv preprint arXiv:2308.12966}, 2023.

\bibitem[Bai et~al.(2025{\natexlab{a}})Bai, Cai, Chen, Chen, Chen, Cheng, Deng,
  Ding, Gao, Ge, Ge, Guo, Huang, Huang, Huang, Hui, Jiang, Li, Li, Li, Li, Lin,
  Lin, Liu, Liu, Liu, Liu, Liu, Liu, Lu, Luo, Lv, Men, Meng, Ren, Ren, Song,
  Sun, Tang, Tu, Wan, Wang, Wang, Wang, Wang, Xie, Xu, Xu, Xu, Yang, Yang,
  Yang, Yang, Yu, Zhang, Zhang, Zhang, Zheng, Zhong, Zhou, Zhou, Zhou, Zhu, and
  Zhu]{Qwen3-VL}
Shuai Bai, Yuxuan Cai, Ruizhe Chen, Keqin Chen, Xionghui Chen, Zesen Cheng,
  Lianghao Deng, Wei Ding, Chang Gao, Chunjiang Ge, Wenbin Ge, Zhifang Guo,
  Qidong Huang, Jie Huang, Fei Huang, Binyuan Hui, Shutong Jiang, Zhaohai Li,
  Mingsheng Li, Mei Li, Kaixin Li, Zicheng Lin, Junyang Lin, Xuejing Liu,
  Jiawei Liu, Chenglong Liu, Yang Liu, Dayiheng Liu, Shixuan Liu, Dunjie Lu,
  Ruilin Luo, Chenxu Lv, Rui Men, Lingchen Meng, Xuancheng Ren, Xingzhang Ren,
  Sibo Song, Yuchong Sun, Jun Tang, Jianhong Tu, Jianqiang Wan, Peng Wang,
  Pengfei Wang, Qiuyue Wang, Yuxuan Wang, Tianbao Xie, Yiheng Xu, Haiyang Xu,
  Jin Xu, Zhibo Yang, Mingkun Yang, Jianxin Yang, An Yang, Bowen Yu, Fei Zhang,
  Hang Zhang, Xi Zhang, Bo Zheng, Humen Zhong, Jingren Zhou, Fan Zhou, Jing
  Zhou, Yuanzhi Zhu, and Ke Zhu.
\newblock Qwen3-vl technical report.
\newblock \emph{arXiv preprint arXiv:2511.21631}, 2025{\natexlab{a}}.

\bibitem[Bai et~al.(2025{\natexlab{b}})Bai, Chen, Liu, Wang, Ge, Song, Dang,
  Wang, Wang, Tang, Zhong, Zhu, Yang, Li, Wan, Wang, Ding, Fu, Xu, Ye, Zhang,
  Xie, Cheng, Zhang, Yang, Xu, and Lin]{Qwen2.5-VL}
Shuai Bai, Keqin Chen, Xuejing Liu, Jialin Wang, Wenbin Ge, Sibo Song, Kai
  Dang, Peng Wang, Shijie Wang, Jun Tang, Humen Zhong, Yuanzhi Zhu, Mingkun
  Yang, Zhaohai Li, Jianqiang Wan, Pengfei Wang, Wei Ding, Zheren Fu, Yiheng
  Xu, Jiabo Ye, Xi Zhang, Tianbao Xie, Zesen Cheng, Hang Zhang, Zhibo Yang,
  Haiyang Xu, and Junyang Lin.
\newblock Qwen2.5-vl technical report.
\newblock \emph{arXiv preprint arXiv:2502.13923}, 2025{\natexlab{b}}.

\bibitem[Chen et~al.(2025)Chen, Razin, Narasimhan, and Chen]{chen2510retaining}
Howard Chen, Noam Razin, Karthik Narasimhan, and Danqi Chen.
\newblock Retaining by doing: The role of on-policy data in mitigating
  forgetting, 2025.
\newblock \emph{URL https://arxiv. org/abs/2510.18874}, 2025.

\bibitem[Cheng et~al.(2024)Cheng, Sun, Chu, Xu, YanTao, Zhang, and
  Wu]{cheng2024seeclick}
Kanzhi Cheng, Qiushi Sun, Yougang Chu, Fangzhi Xu, Li YanTao, Jianbing Zhang,
  and Zhiyong Wu.
\newblock {S}ee{C}lick: Harnessing {GUI} grounding for advanced visual {GUI}
  agents.
\newblock In \emph{Proceedings of the 62nd Annual Meeting of the Association
  for Computational Linguistics (Volume 1: Long Papers)}, pages 9313--9332,
  Bangkok, Thailand, 2024. Association for Computational Linguistics.

\bibitem[Deng et~al.(2023)Deng, Gu, Zheng, Chen, Stevens, Wang, Sun, and
  Su]{deng2023mind2web}
Xiang Deng, Yu Gu, Boyuan Zheng, Shijie Chen, Sam Stevens, Boshi Wang, Huan
  Sun, and Yu Su.
\newblock Mind2web: Towards a generalist agent for the web.
\newblock \emph{Advances in Neural Information Processing Systems},
  36:\penalty0 28091--28114, 2023.

\bibitem[Gao et~al.(2025)Gao, Zheng, Chen, Dang, Liu, Yu, Yang, Bai, Zhou, and
  Lin]{gao2025softadaptivepolicyoptimization}
Chang Gao, Chujie Zheng, Xiong-Hui Chen, Kai Dang, Shixuan Liu, Bowen Yu, An
  Yang, Shuai Bai, Jingren Zhou, and Junyang Lin.
\newblock Soft adaptive policy optimization, 2025.

\bibitem[Gou et~al.(2025)Gou, Wang, Zheng, Xie, Chang, Shu, Sun, and
  Su]{gou2025uground}
Boyu Gou, Ruohan Wang, Boyuan Zheng, Yanan Xie, Cheng Chang, Yiheng Shu, Huan
  Sun, and Yu Su.
\newblock Navigating the digital world as humans do: Universal visual grounding
  for {GUI} agents.
\newblock In \emph{The Thirteenth International Conference on Learning
  Representations}, 2025.

\bibitem[Guo et~al.(2025)Guo, Wu, Zhu, Leng, Shi, Chen, Fan, Wang, Jiang, Wang,
  Chen, Huang, Lei, Yuan, Luo, Liu, Ye, Qian, Yan, Zhao, Peng, Li, Yuan, Wu,
  Cheng, Liu, Wang, Zeng, Liu, Qin, Ding, Xiao, Zhang, Zhang, Xiong, Peng,
  Chen, Li, Hu, Lin, Hu, Zhang, Wu, Li, Liu, Ling, Qin, Wang, He, Zhang, Yi,
  Liao, Huang, Zhang, Deng, Deng, Lin, Yuan, Li, Gou, Lou, Wei, Liu, Li, Zhu,
  Zhong, Li, Zhang, Wu, Li, Xiao, Lin, Yang, Wang, Ji, Hao, Shen, Li, Li, Wu,
  Zhu, Jiao, Feng, Chen, Duan, Liu, Zeng, Tang, Sun, Chen, Long, Feng, Zhan,
  Fang, Lu, Hua, Liu, Shen, Zhang, Shen, Wang, Pan, Zhang, Li, Li, Li, Shi,
  Han, Xiang, Chen, Chen, Li, Yan, Chi, Liu, Du, Wang, Pan, Chen, Chen, Wu,
  Yuan, Shuai, Tao, Zheng, Zhang, Zhang, Wang, Yang, Zhao, Xu, Liang, Yan,
  Zhong, Cao, Wu, Liu, Chang, Cai, Ao, Yang, Zhang, Zhong, Jia, Weng, Yu,
  Huang, Zhu, Yang, Wang, Long, Yin, Li, Zhu, Jia, Zhang, Liu, Zhang, Yang,
  Luo, Chen, Zhong, Xiao, Li, Wu, Wen, Du, Zhang, Ye, Wu, Liu, Yue, Zhou, Yuan,
  Xu, Yang, Zhang, Fang, Li, Ren, Xiong, Hong, Wang, Sun, Wang, Cai, Zha, An,
  Zhao, Xu, Chen, Wu, Zheng, Wang, Huang, Zhu, and
  Song]{guo2025seed15vltechnicalreport}
Dong Guo, Faming Wu, Feida Zhu, Fuxing Leng, Guang Shi, Haobin Chen, Haoqi Fan,
  Jian Wang, Jianyu Jiang, Jiawei Wang, Jingji Chen, Jingjia Huang, Kang Lei,
  Liping Yuan, Lishu Luo, Pengfei Liu, Qinghao Ye, Rui Qian, Shen Yan, Shixiong
  Zhao, Shuai Peng, Shuangye Li, Sihang Yuan, Sijin Wu, Tianheng Cheng, Weiwei
  Liu, Wenqian Wang, Xianhan Zeng, Xiao Liu, Xiaobo Qin, Xiaohan Ding, Xiaojun
  Xiao, Xiaoying Zhang, Xuanwei Zhang, Xuehan Xiong, Yanghua Peng, Yangrui
  Chen, Yanwei Li, Yanxu Hu, Yi Lin, Yiyuan Hu, Yiyuan Zhang, Youbin Wu, Yu Li,
  Yudong Liu, Yue Ling, Yujia Qin, Zanbo Wang, Zhiwu He, Aoxue Zhang, Bairen
  Yi, Bencheng Liao, Can Huang, Can Zhang, Chaorui Deng, Chaoyi Deng, Cheng
  Lin, Cheng Yuan, Chenggang Li, Chenhui Gou, Chenwei Lou, Chengzhi Wei,
  Chundian Liu, Chunyuan Li, Deyao Zhu, Donghong Zhong, Feng Li, Feng Zhang,
  Gang Wu, Guodong Li, Guohong Xiao, Haibin Lin, Haihua Yang, Haoming Wang,
  Heng Ji, Hongxiang Hao, Hui Shen, Huixia Li, Jiahao Li, Jialong Wu, Jianhua
  Zhu, Jianpeng Jiao, Jiashi Feng, Jiaze Chen, Jianhui Duan, Jihao Liu, Jin
  Zeng, Jingqun Tang, Jingyu Sun, Joya Chen, Jun Long, Junda Feng, Junfeng
  Zhan, Junjie Fang, Junting Lu, Kai Hua, Kai Liu, Kai Shen, Kaiyuan Zhang, Ke
  Shen, Ke Wang, Keyu Pan, Kun Zhang, Kunchang Li, Lanxin Li, Lei Li, Lei Shi,
  Li Han, Liang Xiang, Liangqiang Chen, Lin Chen, Lin Li, Lin Yan, Liying Chi,
  Longxiang Liu, Mengfei Du, Mingxuan Wang, Ningxin Pan, Peibin Chen, Pengfei
  Chen, Pengfei Wu, Qingqing Yuan, Qingyao Shuai, Qiuyan Tao, Renjie Zheng,
  Renrui Zhang, Ru Zhang, Rui Wang, Rui Yang, Rui Zhao, Shaoqiang Xu, Shihao
  Liang, Shipeng Yan, Shu Zhong, Shuaishuai Cao, Shuangzhi Wu, Shufan Liu,
  Shuhan Chang, Songhua Cai, Tenglong Ao, Tianhao Yang, Tingting Zhang, Wanjun
  Zhong, Wei Jia, Wei Weng, Weihao Yu, Wenhao Huang, Wenjia Zhu, Wenli Yang,
  Wenzhi Wang, Xiang Long, XiangRui Yin, Xiao Li, Xiaolei Zhu, Xiaoying Jia,
  Xijin Zhang, Xin Liu, Xinchen Zhang, Xinyu Yang, Xiongcai Luo, Xiuli Chen,
  Xuantong Zhong, Xuefeng Xiao, Xujing Li, Yan Wu, Yawei Wen, Yifan Du, Yihao
  Zhang, Yining Ye, Yonghui Wu, Yu Liu, Yu Yue, Yufeng Zhou, Yufeng Yuan,
  Yuhang Xu, Yuhong Yang, Yun Zhang, Yunhao Fang, Yuntao Li, Yurui Ren, Yuwen
  Xiong, Zehua Hong, Zehua Wang, Zewei Sun, Zeyu Wang, Zhao Cai, Zhaoyue Zha,
  Zhecheng An, Zhehui Zhao, Zhengzhuo Xu, Zhipeng Chen, Zhiyong Wu, Zhuofan
  Zheng, Zihao Wang, Zilong Huang, Ziyu Zhu, and Zuquan Song.
\newblock Seed1.5-vl technical report, 2025.

\bibitem[Hong et~al.(2024)Hong, Wang, Lv, Xu, Yu, Ji, Wang, Wang, Dong, Ding,
  et~al.]{hong2024cogagent}
Wenyi Hong, Weihan Wang, Qingsong Lv, Jiazheng Xu, Wenmeng Yu, Junhui Ji, Yan
  Wang, Zihan Wang, Yuxiao Dong, Ming Ding, et~al.
\newblock Cogagent: A visual language model for gui agents.
\newblock In \emph{Proceedings of the IEEE/CVF conference on computer vision
  and pattern recognition}, pages 14281--14290, 2024.

\bibitem[Hsieh et~al.(2025)Hsieh, Wei, and Yang]{hsieh2025zonui3b}
ZongHan Hsieh, Tzer-Jen Wei, and ShengJing Yang.
\newblock Zonui-3b: A lightweight vision-language model for cross-resolution
  gui grounding.
\newblock \url{https://arxiv.org/abs/2506.23491}, 2025.
\newblock arXiv:2506.23491 [cs.CV], version 2, last revised 1 Jul 2025.

\bibitem[Huang et~al.(2025)Huang, Yu, Ma, Zhong, Feng, Wang, Chen, Peng, Feng,
  Qin, et~al.]{huang2025survey}
Lei Huang, Weijiang Yu, Weitao Ma, Weihong Zhong, Zhangyin Feng, Haotian Wang,
  Qianglong Chen, Weihua Peng, Xiaocheng Feng, Bing Qin, et~al.
\newblock A survey on hallucination in large language models: Principles,
  taxonomy, challenges, and open questions.
\newblock \emph{ACM Transactions on Information Systems}, 43\penalty0
  (2):\penalty0 1--55, 2025.

\bibitem[Hurst et~al.(2024)Hurst, Lerer, Goucher, Perelman, Ramesh, Clark,
  Ostrow, Welihinda, Hayes, Radford, et~al.]{hurst2024gpt}
Aaron Hurst, Adam Lerer, Adam~P Goucher, Adam Perelman, Aditya Ramesh, Aidan
  Clark, AJ Ostrow, Akila Welihinda, Alan Hayes, Alec Radford, et~al.
\newblock Gpt-4o system card.
\newblock \emph{arXiv preprint arXiv:2410.21276}, 2024.

\bibitem[Li et~al.(2025)Li, Ziyang, Lin, Luo, Tian, Ma, Huang, and
  Chua]{li2025screenspotpro}
Kaixin Li, Meng Ziyang, Hongzhan Lin, Ziyang Luo, Yuchen Tian, Jing Ma, Zhiyong
  Huang, and Tat-Seng Chua.
\newblock Screenspot-pro: {GUI} grounding for professional high-resolution
  computer use.
\newblock In \emph{Workshop on Reasoning and Planning for Large Language
  Models}, 2025.

\bibitem[Lin et~al.(2024)Lin, Li, Gao, Yang, Wu, Bai, Lei, Wang, and
  Shou]{lin2024showui}
Kevin~Qinghong Lin, Linjie Li, Difei Gao, Zhengyuan Yang, Shiwei Wu, Zechen
  Bai, Weixian Lei, Lijuan Wang, and Mike~Zheng Shou.
\newblock Showui: One vision-language-action model for gui visual agent, 2024.

\bibitem[Liu et~al.(2025{\natexlab{a}})Liu, Zhang, Xu, Wanyan, Wang, Yan,
  Zhang, Yuan, Xu, Hu, et~al.]{liu2025pc}
Haowei Liu, Xi Zhang, Haiyang Xu, Yuyang Wanyan, Junyang Wang, Ming Yan, Ji
  Zhang, Chunfeng Yuan, Changsheng Xu, Weiming Hu, et~al.
\newblock Pc-agent: A hierarchical multi-agent collaboration framework for
  complex task automation on pc.
\newblock \emph{arXiv preprint arXiv:2502.14282}, 2025{\natexlab{a}}.

\bibitem[Liu et~al.(2025{\natexlab{b}})Liu, Li, Xie, Hu, Han, Zhang, Yang, and
  Wu]{liu2025infigui}
Yuhang Liu, Pengxiang Li, Congkai Xie, Xavier Hu, Xiaotian Han, Shengyu Zhang,
  Hongxia Yang, and Fei Wu.
\newblock Infigui-r1: Advancing multimodal gui agents from reactive actors to
  deliberative reasoners.
\newblock \emph{arXiv preprint arXiv:2504.14239}, 2025{\natexlab{b}}.

\bibitem[Lopez-Paz et~al.(2015)Lopez-Paz, Bottou, Sch{\"o}lkopf, and
  Vapnik]{lopez2015unifying}
David Lopez-Paz, L{\'e}on Bottou, Bernhard Sch{\"o}lkopf, and Vladimir Vapnik.
\newblock Unifying distillation and privileged information.
\newblock \emph{arXiv preprint arXiv:1511.03643}, 2015.

\bibitem[Lu et~al.(2025)Lu, Chai, Guo, Yin, Liu, Wang, Xiong, and Li]{lu2025ui}
Zhengxi Lu, Yuxiang Chai, Yaxuan Guo, Xi Yin, Liang Liu, Hao Wang, Guanjing
  Xiong, and Hongsheng Li.
\newblock Ui-r1: Enhancing action prediction of gui agents by reinforcement
  learning.
\newblock \emph{arXiv preprint arXiv:2503.21620}, 2025.

\bibitem[Luo et~al.(2025)Luo, Wang, He, and Xia]{luo2025gui}
Run Luo, Lu Wang, Wanwei He, and Xiaobo Xia.
\newblock Gui-r1: A generalist r1-style vision-language action model for gui
  agents.
\newblock \emph{arXiv preprint arXiv:2504.10458}, 2025.

\bibitem[openai(2025)]{gpt_o3}
openai.
\newblock Openai o3.
\newblock \url{https://openai.com}, 2025.

\bibitem[Parker~Lyman(2025)]{manus}
Tao~Zhang Parker~Lyman.
\newblock Manus academy.
\newblock \url{https://manus.im}, 2025.

\bibitem[Qin et~al.(2025)Qin, Ye, Fang, Wang, Liang, Tian, Zhang, Li, Li,
  Huang, et~al.]{qin2025ui}
Yujia Qin, Yining Ye, Junjie Fang, Haoming Wang, Shihao Liang, Shizuo Tian,
  Junda Zhang, Jiahao Li, Yunxin Li, Shijue Huang, et~al.
\newblock Ui-tars: Pioneering automated gui interaction with native agents.
\newblock \emph{arXiv preprint arXiv:2501.12326}, 2025.

\bibitem[Ross et~al.(2011)Ross, Gordon, and Bagnell]{ross2011reduction}
St{\'e}phane Ross, Geoffrey Gordon, and Drew Bagnell.
\newblock A reduction of imitation learning and structured prediction to
  no-regret online learning.
\newblock In \emph{Proceedings of the fourteenth international conference on
  artificial intelligence and statistics}, pages 627--635. JMLR Workshop and
  Conference Proceedings, 2011.

\bibitem[Shao et~al.(2024)Shao, Wang, Zhu, Xu, Song, Bi, Zhang, Zhang, Li, Wu,
  and Guo]{shao2024deepseekmathpushinglimitsmathematical}
Zhihong Shao, Peiyi Wang, Qihao Zhu, Runxin Xu, Junxiao Song, Xiao Bi, Haowei
  Zhang, Mingchuan Zhang, Y.~K. Li, Y. Wu, and Daya Guo.
\newblock Deepseekmath: Pushing the limits of mathematical reasoning in open
  language models, 2024.

\bibitem[Shen et~al.(2025)Shen, Liu, Li, Fang, Ma, Liao, Shen, Zhang, Zhao,
  Zhang, Xu, and Zhao]{shen2025vlm}
Haozhan Shen, Peng Liu, Jingcheng Li, Chunxin Fang, Yibo Ma, Jiajia Liao,
  Qiaoli Shen, Zilun Zhang, Kangjia Zhao, Qianqian Zhang, Ruochen Xu, and
  Tiancheng Zhao.
\newblock Vlm-r1: A stable and generalizable r1-style large vision-language
  model.
\newblock \emph{arXiv preprint arXiv:2504.07615}, 2025.

\bibitem[Shenfeld et~al.(2026)Shenfeld, Damani, H{\"u}botter, and
  Agrawal]{shenfeld2026self}
Idan Shenfeld, Mehul Damani, Jonas H{\"u}botter, and Pulkit Agrawal.
\newblock Self-distillation enables continual learning.
\newblock \emph{arXiv preprint arXiv:2601.19897}, 2026.

\bibitem[Team et~al.(2023)Team, Anil, Borgeaud, Alayrac, Yu, Soricut,
  Schalkwyk, Dai, Hauth, Millican, et~al.]{team2023gemini}
Gemini Team, Rohan Anil, Sebastian Borgeaud, Jean-Baptiste Alayrac, Jiahui Yu,
  Radu Soricut, Johan Schalkwyk, Andrew~M Dai, Anja Hauth, Katie Millican,
  et~al.
\newblock Gemini: a family of highly capable multimodal models.
\newblock \emph{arXiv preprint arXiv:2312.11805}, 2023.

\bibitem[Team et~al.(2025)Team, Du, Yin, Xing, Qu, Wang, Chen, Zhang, Du, Wei,
  Wang, Zhang, Du, Wang, Yuan, Lu, Li, Sung, Wei, Lai, Zhu, Ding, Hu, Yang,
  Zhang, Wu, Yao, Lu, Wang, Gao, Zheng, Li, Su, Wang, Deng, Qiu, Xie, Wang,
  Liu, Yan, Ouyang, Chen, Sui, Yu, Dong, Dong, Xu, Cheng, Gu, Zhou, Liu, Cao,
  Yu, Song, Bai, Song, He, Huang, Xu, Yuan, Yao, Wu, Zu, Zhou, Wang, Charles,
  Zhong, Li, Hu, Chen, Wang, Liu, Miao, Qin, Chen, Bao, Wang, Kang, Liu, Du,
  Wu, Wang, Yan, Zhou, Li, Jiang, Zhang, Yang, Huang, Huang, Zhao, and
  Chen]{kimiteam2025kimivltechnicalreport}
Kimi Team, Angang Du, Bohong Yin, Bowei Xing, Bowen Qu, Bowen Wang, Cheng Chen,
  Chenlin Zhang, Chenzhuang Du, Chu Wei, Congcong Wang, Dehao Zhang, Dikang Du,
  Dongliang Wang, Enming Yuan, Enzhe Lu, Fang Li, Flood Sung, Guangda Wei,
  Guokun Lai, Han Zhu, Hao Ding, Hao Hu, Hao Yang, Hao Zhang, Haoning Wu,
  Haotian Yao, Haoyu Lu, Heng Wang, Hongcheng Gao, Huabin Zheng, Jiaming Li,
  Jianlin Su, Jianzhou Wang, Jiaqi Deng, Jiezhong Qiu, Jin Xie, Jinhong Wang,
  Jingyuan Liu, Junjie Yan, Kun Ouyang, Liang Chen, Lin Sui, Longhui Yu,
  Mengfan Dong, Mengnan Dong, Nuo Xu, Pengyu Cheng, Qizheng Gu, Runjie Zhou,
  Shaowei Liu, Sihan Cao, Tao Yu, Tianhui Song, Tongtong Bai, Wei Song, Weiran
  He, Weixiao Huang, Weixin Xu, Xiaokun Yuan, Xingcheng Yao, Xingzhe Wu,
  Xinxing Zu, Xinyu Zhou, Xinyuan Wang, Y. Charles, Yan Zhong, Yang Li,
  Yangyang Hu, Yanru Chen, Yejie Wang, Yibo Liu, Yibo Miao, Yidao Qin, Yimin
  Chen, Yiping Bao, Yiqin Wang, Yongsheng Kang, Yuanxin Liu, Yulun Du, Yuxin
  Wu, Yuzhi Wang, Yuzi Yan, Zaida Zhou, Zhaowei Li, Zhejun Jiang, Zheng Zhang,
  Zhilin Yang, Zhiqi Huang, Zihao Huang, Zijia Zhao, and Ziwei Chen.
\newblock {Kimi-VL} technical report, 2025.

\bibitem[Vapnik and Izmailov(2015)]{vapnik2015learning}
Vladimir Vapnik and Rauf Izmailov.
\newblock Learning using privileged information: similarity control and
  knowledge transfer.
\newblock \emph{The Journal of Machine Learning Research}, 16\penalty0
  (1):\penalty0 2023--2049, 2015.

\bibitem[Vapnik and Vashist(2009)]{vapnik2009new}
Vladimir Vapnik and Akshay Vashist.
\newblock A new learning paradigm: Learning using privileged information.
\newblock \emph{Neural networks}, 22\penalty0 (5-6):\penalty0 544--557, 2009.

\bibitem[Wang et~al.(2025)Wang, Wang, Lu, Yang, Xie, Wang, Deng, Guo, Xu, Wu,
  Shen, Li, Li, Li, Chen, Zheng, Li, Lei, Cao, Fu, Shin, Shin, Hu, Wang, Chen,
  Ye, Zhang, Du, Hu, Chen, Zhou, Yao, Chen, Gu, Wang, Wang, Yang, Zhong, Sung,
  Charles, Yang, and Yu]{wang2025opencuaopenfoundationscomputeruse}
Xinyuan Wang, Bowen Wang, Dunjie Lu, Junlin Yang, Tianbao Xie, Junli Wang,
  Jiaqi Deng, Xiaole Guo, Yiheng Xu, Chen~Henry Wu, Zhennan Shen, Zhuokai Li,
  Ryan Li, Xiaochuan Li, Junda Chen, Boyuan Zheng, Peihang Li, Fangyu Lei,
  Ruisheng Cao, Yeqiao Fu, Dongchan Shin, Martin Shin, Jiarui Hu, Yuyan Wang,
  Jixuan Chen, Yuxiao Ye, Danyang Zhang, Dikang Du, Hao Hu, Huarong Chen, Zaida
  Zhou, Haotian Yao, Ziwei Chen, Qizheng Gu, Yipu Wang, Heng Wang, Diyi Yang,
  Victor Zhong, Flood Sung, Y. Charles, Zhilin Yang, and Tao Yu.
\newblock Opencua: Open foundations for computer-use agents, 2025.

\bibitem[Wu et~al.(2025)Wu, Cheng, Yang, Zhang, Yang, Jiang, Mu, Peng, Qiao,
  Tan, et~al.]{wu2025gui}
Qianhui Wu, Kanzhi Cheng, Rui Yang, Chaoyun Zhang, Jianwei Yang, Huiqiang
  Jiang, Jian Mu, Baolin Peng, Bo Qiao, Reuben Tan, et~al.
\newblock Gui-actor: Coordinate-free visual grounding for gui agents.
\newblock \emph{arXiv preprint arXiv:2506.03143}, 2025.

\bibitem[Wu et~al.(2024)Wu, Wu, Xu, Wang, Sun, Jia, Cheng, Ding, Chen, Liang,
  et~al.]{wu2024atlas}
Zhiyong Wu, Zhenyu Wu, Fangzhi Xu, Yian Wang, Qiushi Sun, Chengyou Jia, Kanzhi
  Cheng, Zichen Ding, Liheng Chen, Paul~Pu Liang, et~al.
\newblock Os-atlas: A foundation action model for generalist gui agents.
\newblock \emph{arXiv preprint arXiv:2410.23218}, 2024.

\bibitem[Xie et~al.(2024)Xie, Zhang, Chen, Li, Zhao, Cao, Hua, Cheng, Shin,
  Lei, Liu, Xu, Zhou, Savarese, Xiong, Zhong, and Yu]{OSWorld}
Tianbao Xie, Danyang Zhang, Jixuan Chen, Xiaochuan Li, Siheng Zhao, Ruisheng
  Cao, Toh~Jing Hua, Zhoujun Cheng, Dongchan Shin, Fangyu Lei, Yitao Liu,
  Yiheng Xu, Shuyan Zhou, Silvio Savarese, Caiming Xiong, Victor Zhong, and Tao
  Yu.
\newblock Osworld: Benchmarking multimodal agents for open-ended tasks in real
  computer environments, 2024.

\bibitem[Xie et~al.(2025)Xie, Deng, Li, Yang, Wu, Chen, Hu, Wang, Xu, Wang, Xu,
  Wang, Sahoo, Yu, and Xiong]{xie2025scalingcomputerusegroundinguser}
Tianbao Xie, Jiaqi Deng, Xiaochuan Li, Junlin Yang, Haoyuan Wu, Jixuan Chen,
  Wenjing Hu, Xinyuan Wang, Yuhui Xu, Zekun Wang, Yiheng Xu, Junli Wang, Doyen
  Sahoo, Tao Yu, and Caiming Xiong.
\newblock Scaling computer-use grounding via user interface decomposition and
  synthesis, 2025.

\bibitem[Xu et~al.(2025)Xu, Yang, Chen, Wang, Chen, and Tang]{xu2025deskvision}
Yibin Xu, Liang Yang, Hao Chen, Hua Wang, Zhi Chen, and Yaohua Tang.
\newblock Deskvision: Large scale desktop region captioning for advanced gui
  agents.
\newblock \emph{arXiv preprint arXiv:2503.11170}, 2025.

\bibitem[Yang et~al.(2024)Yang, Wang, Li, Luo, Chen, Huang, and Li]{ariaui}
Yuhao Yang, Yue Wang, Dongxu Li, Ziyang Luo, Bei Chen, Chao Huang, and Junnan
  Li.
\newblock Aria-ui: Visual grounding for gui instructions.
\newblock \emph{arXiv preprint arXiv:2412.16256}, 2024.

\bibitem[Yuan et~al.(2025)Yuan, Zhang, Li, Cai, Yao, Chen, Wang, Hou, Chen,
  Jiang, et~al.]{yuan2025enhancing}
Xinbin Yuan, Jian Zhang, Kaixin Li, Zhuoxuan Cai, Lujian Yao, Jie Chen, Enguang
  Wang, Qibin Hou, Jinwei Chen, Peng-Tao Jiang, et~al.
\newblock Enhancing visual grounding for gui agents via self-evolutionary
  reinforcement learning.
\newblock \emph{arXiv preprint arXiv:2505.12370}, 2025.

\bibitem[Zeng et~al.(2024)Zeng, Liu, Lu, Wang, Liu, Dong, and
  Tang]{zeng2024agenttuning}
Aohan Zeng, Mingdao Liu, Rui Lu, Bowen Wang, Xiao Liu, Yuxiao Dong, and Jie
  Tang.
\newblock Agenttuning: Enabling generalized agent abilities for llms.
\newblock In \emph{Findings of the Association for Computational Linguistics:
  ACL 2024}, pages 3053--3077, 2024.

\bibitem[Zhou et~al.(2025)Zhou, Dai, Wang, Zhou, Jia, and Xu]{zhou2025guig1}
Yuqi Zhou, Sunhao Dai, Shuai Wang, Kaiwen Zhou, Qinglin Jia, and Jun Xu.
\newblock Gui-g1: Understanding r1-zero-like training for visual grounding in
  gui agents.
\newblock \emph{arXiv preprint arXiv:2505.15810}, 2025.

\end{thebibliography}
}

\clearpage
\appendix

\section{Ground-truth Guidance Variants}
\label{app:gt_guidance}

Let $\mathcal{A}_t^*$ denote the human-verified valid action set for the current GUI state. The teacher-side guidance $g_t$ is constructed from $\mathcal{A}_t^*$.

For Single-GT Guided OPD, one valid action is randomly sampled:
\begin{equation}
g_t = a_t^{*(k)}, \quad k\sim \mathrm{Uniform}\{1,\dots,K\}.
\end{equation}

For Multi-GT Guided OPD, the full valid action set is provided:
\begin{equation}
g_t = \mathcal{A}_t^*.
\end{equation}

For Guided OPD, the guidance is selected according to:
\begin{equation}
g_t = a_t^\dagger =
\arg\max_{a^*\in \mathcal{A}_t^*} S(\hat{a}_t,a^*).
\end{equation}

All three variants use the same student input $x_t$ and differ only in the privileged teacher context.

\section{Action Matching and Multi-solution Reward Details}
\label{app:reward_details}

We define the base GUI action matcher $\phi_{\mathrm{gui}}(\hat{y}_t,a^*)$ as a normalized score in $[0,1]$. If the model output cannot be parsed as a valid JSON action or violates the required action schema, the score is set to zero.
For the reported runs, the component weights are
$w_{\tau}=10$, $w_p=5$, and $w_v=4$; with the value-matching scores below,
an exact value match contributes 4 and a partial match contributes 2 before
normalization.

\paragraph{Action type matching.}
\begin{equation}
m_{\tau}=\mathbb{I}[\hat{\tau}=\tau^*].
\end{equation}

\paragraph{Position matching.}
For position-based actions, the model predicts a point $\hat{p}$ and the annotation provides a target bounding box $b^*$. We use a continuous position score $m_p\in[0,1]$. If the predicted point falls inside the central region of the target box, $m_p=1$. If it falls inside the full box but outside the central region, the score is linearly interpolated between $0.5$ and $1$. If it falls outside the box, the score decays exponentially with the distance $d$ to the box boundary:
\begin{equation}
m_p = 0.5\exp(-d/200).
\end{equation}

\paragraph{Value matching.}
For text or key-value actions, we compute:
\begin{equation}
m_v=
\begin{cases}
1, & \text{exact match},\\
0.5, & \text{partial match},\\
0, & \text{otherwise}.
\end{cases}
\end{equation}

\paragraph{Combined score.}
The final action matching score is:
\begin{equation}
\phi_{\mathrm{gui}}(\hat{y}_t,a^*)
=
\frac{
w_{\tau}m_{\tau}
+
\mathbb{I}[b^*\neq\emptyset]w_pm_p
+
\mathbb{I}[v^*\neq\emptyset]w_vm_v
}{
w_{\tau}
+
\mathbb{I}[b^*\neq\emptyset]w_p
+
\mathbb{I}[v^*\neq\emptyset]w_v
}.
\end{equation}

Given multiple valid actions, we use:
\begin{equation}
R_{\mathrm{ms}}(\hat{y}_t,\mathcal{A}_t^*)
=
\max_{a^*\in\mathcal{A}_t^*}
\phi_{\mathrm{gui}}(\hat{y}_t,a^*).
\end{equation}

\section{Planning Reward Details}
\label{app:planning_reward}

The planning reward is a history-conditioned, per-state plan-quality signal computed for candidate subtask plans at logged GUI states using an external VLM judge. The judge receives the user instruction, available screenshot history, previous model outputs, and the current candidate subtask plan. Candidate actions are not executed for reward computation, and the reward uses neither sampled future states nor terminal task-success feedback; it is therefore a local surrogate for plan quality, not an estimate of long-term return or verified task recovery.

The judge evaluates the plan along four dimensions:
\begin{enumerate}
    \item \textbf{Task relevance and adaptability}: whether the plan reflects the user instruction and adapts to the current UI state.
    \item \textbf{Visual grounding}: whether the plan is grounded in actual UI elements visible in the screenshots.
    \item \textbf{Decomposition granularity}: whether the plan decomposes the task into actionable subtasks at an appropriate granularity.
    \item \textbf{State consistency}: whether the completion status of subtasks is consistent with the latest screenshot.
\end{enumerate}

If the latest screenshot indicates that the agent is stuck, in an error state, or in a loop, the first unfinished subtask must explicitly propose a corrective next step. If the plan ignores the obstacle, remains overly high-level, or contains completion markers inconsistent with the latest screenshot, the judge is instructed to assign a score below $0.5$.

The judge is required to return a structured JSON score:
\begin{verbatim}
{"score": 0.8}
\end{verbatim}

\section{Training-time Planning Judge: Prompt and Reliability}
\label{app:judge_prompt}

\begin{figure*}[t]
  \centering 
    \includegraphics[width=0.95\textwidth]{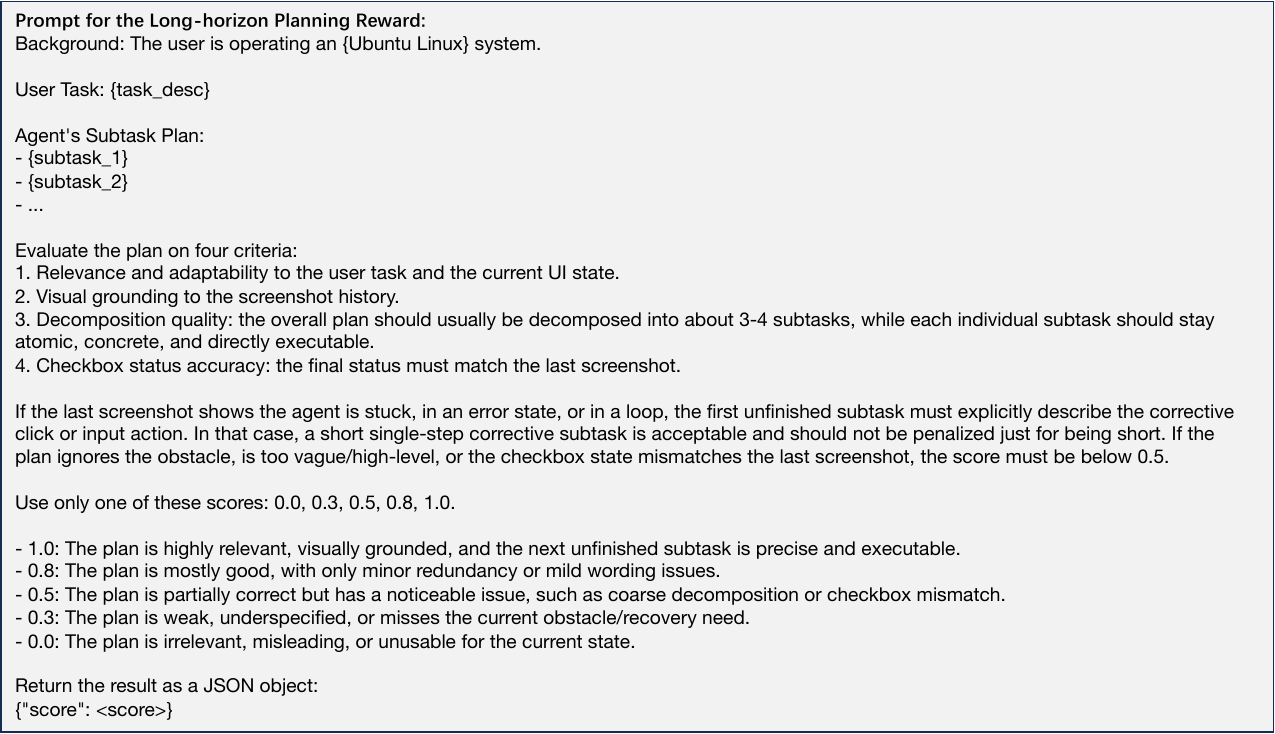}
  \caption{Prompt for the planning reward.}
  \label{fig:long_reward_prompt}
\end{figure*}

We use a frozen Qwen3-VL-32B-Instruct model as the VLM judge for the planning reward. The judge is queried only during the logged-state policy-optimization stage. The temperature is set to $0$, and the judge is instructed to output a structured JSON score from the discrete set $\{0.0, 0.3, 0.5, 0.8, 1.0\}$. For detailed prompt design, please refer to Fig.~ \ref{fig:long_reward_prompt}.

\paragraph{Human--judge agreement.}
On a held-out calibration subset, Qwen3-VL-32B-Instruct achieves
90.0\% exact agreement with human annotations (Cohen's $\kappa=0.7916$).
Treating high-quality
trajectories as the positive class gives 94.52\% recall, 81.18\% precision,
and 87.34\% F1. These metrics assess the training-time planning reward
judge, not the GPT-5.4 evaluator used to judge Lite-Bench task success
(Appendix~\ref{app:litebench}).

\paragraph{Scope and limitations.}
The lower precision suggests an optimistic bias; this
offline agreement study does not rule out reward hacking during training.
The reward coefficients are 0.30 for planning, 0.60 for action matching, and
0.10 for format/length; these are weights, not fixed shares of realized reward.

\section{Generalized Knowledge Distillation}
\label{app:gkd_training}

We first optimize the policy with generalized knowledge distillation (GKD). The teacher is Qwen3-VL-32B-Instruct, and the students are Qwen3-VL-2B-Instruct and Qwen3-VL-30B-A3B. All GKD runs use full-parameter bfloat16 training with the Megatron backend. Unlike offline distillation from fixed target responses, our GKD stage is on-policy at the logged-state sampling level: the student generates the completion that is used for distillation, but the sampled GUI action is not executed in an environment. Concretely, we set the GKD on-policy probability to $\lambda=1.0$, use vLLM for batched completion sampling, and disable sequential teacher generation. The dataset therefore provides prompts and human-verified valid-action candidates for constructing the privileged guidance described in the main text, while the optimized response tokens are sampled from the current student policy.

For each GUI state $x_t$, the student samples a completion $\hat{y}_t \sim \pi_{\theta}(\cdot \mid x_t)$. The sampled completion is held fixed during the optimization step. The teacher and student are each evaluated once on the full sampled completion sequence under teacher forcing. In Guided-OPD, the teacher receives the most-matched valid action $g_t$ from Section~3.2 as privileged guidance, and its input is $\tilde{x}_t=[x_t,\texttt{reference\_action}=g_t]$. The discrete teacher-side guidance $g_t$ is also held fixed during the optimization step. Let $\mathcal{M}_t$ denote the non-masked completion positions in $\hat{y}_t$. This GKD implementation instantiates the Guided-OPD objective in Section~3.2, so we use the same notation as the main text. Using the full-vocabulary next-token distributions at each position $j\in\mathcal{M}_t$, we compute the distillation loss:
\[
\mathcal{L}_{\mathrm{OPD}}
=
\frac{1}{|\mathcal{M}_t|}
\sum_{j \in \mathcal{M}_t}
D_{\beta}\left(
p_{\theta,j}
\;\middle\|\;
\operatorname{sg}\left[p_{T,j}^{g_t}\right]
\right),
\qquad \beta=1,
\]
where $p_{\theta,j}=\pi_{\theta}(\cdot\mid x_t,\hat{y}_{t,<j})$ is the student distribution at position $j$, and $p_{T,j}^{g_t}=\pi_T(\cdot\mid\tilde{x}_t,\hat{y}_{t,<j})$ is the teacher distribution conditioned on the most-matched guidance $g_t$. The operator $\operatorname{sg}[\cdot]$ denotes stop-gradient. Our implementation parameterizes the generalized JSD/KL family by $\beta$; with $\beta=1$, it selects the reverse-KL direction:
\[
D_1(p\|q)=D_{\mathrm{KL}}(p\|q)=\sum_v p(v)\log\frac{p(v)}{q(v)}.
\]
Therefore, the reported experiments use full-vocabulary, per-prefix reverse KL, not a sampled-token surrogate. The teacher forward pass is executed under \texttt{no\_grad} and its logits are detached; gradients flow only through the student distribution. One teacher-forced forward pass over the complete completion provides teacher distributions for all valid prefixes. We use sampling temperature $1.0$ and top-$p=0.98$ for student rollouts.

\begin{table}[t]
\centering
\scriptsize
\resizebox{\columnwidth}{!}{%
\begin{tabular}{lcc}
\hline
 & \textbf{2B GKD} & \textbf{A3B GKD} \\
\hline
Student & Qwen3-VL-2B-Instruct & Qwen3-VL-30B-A3B \\
Teacher & Qwen3-VL-32B-Instruct & Qwen3-VL-32B-Instruct \\
Training GPUs & 1 node $\times$ 8 A100 GPUs& 2 nodes $\times$ 8 A100 GPUs\\
Rollout mode & colocated vLLM & colocated vLLM \\
On-policy probability $\lambda$ & 1.0 & 1.0 \\
GKD divergence parameter $\beta$ & 1.0 & 1.0 \\
Teacher demo mode & GT guided & GT guided \\
Max sequence length & 8192 & 8192 \\
Max completion length & 512 & 512 \\
Micro batch size & 1 & 1 \\
Global batch size & 32 & 32 \\
Epochs & 2 & 2 \\
TP / PP / EP / CP & 1 / 4 / 1 / 1 & 1 / 4 / 2 / 1 \\
vLLM tensor parallel size & 4 & 4 \\
Learning rate & $2.5\times10^{-7}$ & $2.5\times10^{-7}$ \\
Warmup fraction & 0.01 & 0.01 \\
\hline
\end{tabular}%
}
\caption{Main hyperparameters for the GKD stage. TP, PP, EP, and CP denote tensor, pipeline, expert, and context parallelism.}
\label{tab:gkd_hparams}
\end{table}

\section{Logged-State Policy Optimization with GRPO}
\label{app:grpo_training}

After GKD, we further optimize the policy with group relative policy
optimization (GRPO). For each logged prompt, the behavior policy samples
$G=16$ completions; candidates are scored at the same logged GUI state rather
than by executing live environment transitions. Rewards are normalized within
the group to form relative advantages:
\[
\hat{A}_t^{(i)}
=
\frac{R_t^{(i)}-\operatorname{mean}(\{R_t^{(j)}\}_{j=1}^{G})}
{\operatorname{std}(\{R_t^{(j)}\}_{j=1}^{G})+\epsilon_{\mathrm{adv}}}.
\]
Here $\epsilon_{\mathrm{adv}}$ is a numerical stabilizer distinct from the
clipping bounds below. We use sequence-level importance sampling and the
clipped GRPO objective:
\begin{equation*}
\resizebox{\columnwidth}{!}{$
\mathcal{L}_{\mathrm{GRPO}}
=
-\frac{1}{G}
\sum_{i=1}^{G}
\min\!\left(
\rho_t^{(i)} \hat{A}_t^{(i)},\,
\mathrm{clip}(\rho_t^{(i)}, 1{-}\epsilon_{\mathrm{low}}, 1{+}\epsilon_{\mathrm{high}}) \hat{A}_t^{(i)}
\right)
+ \beta_{\mathrm{KL}} D_{\mathrm{KL}}(\pi_\theta\|\pi_{\mathrm{ref}}),
$}
\end{equation*}
where the sequence-level importance ratio is computed from the mean token
log-probability difference:
\begin{equation*}
\resizebox{\columnwidth}{!}{$
\rho_t^{(i)} =
\exp\!\left[
\frac{1}{|\mathcal{M}^{(i)}_t|}
\sum_{j\in\mathcal{M}^{(i)}_t}
\left(
\log \pi_\theta(\hat{y}^{(i)}_{t,j}\mid x_t,\hat{y}^{(i)}_{t,<j})
-\log \pi_{\theta_{\mathrm{old}}}(\hat{y}^{(i)}_{t,j}\mid x_t,\hat{y}^{(i)}_{t,<j})
\right)
\right].
$}
\end{equation*}
Here $\mathcal{M}^{(i)}_t$ denotes the valid completion-token positions and
$\hat{A}_t^{(i)}$ is the group-normalized advantage defined in
Section~3.3. The resulting scalar ratio is shared
by the whole completion and is the only quantity clipped; no token-wise
clipping is applied. We use $\epsilon_{\mathrm{low}}=0.2$,
$\epsilon_{\mathrm{high}}=0.28$, dynamic sampling, and overlong-response
filtering in both 2B and A3B GRPO runs. The KL coefficient is
$\beta_{\mathrm{KL}}=0$ for both runs.

The reward profile is \texttt{local\_subtask}. It combines multi-solution GUI
action matching, subtask quality judged by a local Qwen3-VL-32B-Instruct reward
model, and a JSON-aware soft overlength penalty. The reward functions and
weights passed to training are:
\[
R
=
0.60\,R_{\mathrm{ms}}
+
0.30\,R_{\mathrm{sub}}
+
0.10\,R_{\mathrm{format/length}}.
\]
The subtask reward model is served locally with vLLM. For the 2B run, two local reward servers are used; for the A3B run, four local reward servers are used. For the A3B policy, generated outputs are additionally constrained by a structured JSON regular expression requiring fields for reasoning, subtask decomposition, and the next action.
The $R_{\mathrm{format/length}}$ term is an auxiliary output-validity
regularizer, not a semantic task-success reward: malformed or schema-invalid
JSON receives no format credit, while valid but overlong responses receive a
soft length penalty under the completion limits listed in Table~\ref{tab:grpo_hparams}.

For the controlled ablations in Table~\ref{tab:mul_sub}, all methods use the same collected-data setting: the loader exposes approximately 70K windowed training samples, and the Guided-OPD variants use a fixed 3K subset carrying multi-solution annotations. No new task instances are introduced for these comparisons.
The headline ScreenSpot-Pro result in Table~\ref{tab:screenspot_pro} uses ShowUI-Desktop-8K~\citep{lin2024showui} augmented with multi-solution annotations, whereas the ScreenSpot-Pro value in Table~\ref{tab:mul_sub} uses this common Lite-Dataset setting; consequently, the 46.86\% and 43.13\% values correspond to different training-data protocols rather than a reporting inconsistency, and should not be interpreted as results from the same checkpoint.

\begin{table}[t]
\centering
\scriptsize
\resizebox{\columnwidth}{!}{%
\begin{tabular}{lcc}
\hline
 & \textbf{2B GRPO} & \textbf{A3B GRPO} \\
\hline
Initialization & 2B GKD checkpoint & A3B GKD checkpoint \\
Training GPUs & 2 nodes $\times$ 8  A100 GPUs& 2 nodes $\times$ 8 A100 GPUs\\
Policy rollout mode & colocated vLLM & colocated vLLM \\
Reward model & Qwen3-VL-32B-Instruct & Qwen3-VL-32B-Instruct \\
Reward servers & 2 & 4 \\
Max sequence length & 10240& 10240 \\
Max completion length & 512 & 640 \\
Soft max length & -- & 512 \\
Soft cache length & 128 & 384 \\
Micro batch size & 1 & 1 \\
Global batch size & 64 & 32 \\
Epochs & 1& 1 \\
Generations per prompt & 16 & 16 \\
Steps per generation & 1 & 1\\
TP / PP / EP / CP & 1 / 4 / 1 / 1 & 1 / 4 / 2 / 1 \\
Sequence parallel & enabled & enabled \\
KL coefficient $\beta_{\mathrm{KL}}$ & 0.0 & 0.0\\
Learning rate & $2.5\times10^{-7}$ & $1.0\times10^{-7}$ \\
Warmup fraction & 0.01 & 0.01 \\
\hline
\end{tabular}%
}
\caption{Main hyperparameters for the GRPO stage. The 2B GRPO run uses a two-stage schedule in which the second stage resumes from the latest checkpoint of the first stage. Both runs use $G=16$, $\epsilon_{\mathrm{low}}=0.2$, $\epsilon_{\mathrm{high}}=0.28$, $\beta_{\mathrm{KL}}=0$, dynamic sampling, and overlong-response filtering.}
\label{tab:grpo_hparams}
\end{table}

\section{Data Pipeline}
\label{app:data_pipeline}

To facilitate the training of LiteGUI, we developed a two-stage data construction pipeline consisting of an automated trajectory generation phase and a high-quality multi-solution annotation phase, as illustrated in Figure~\ref{fig:acc}. Both Lite-Dataset and Lite-Bench are generated through this pipeline.

\subsection{Automated Trajectory Generation (Data Pipeline)}
In this section, we detail the \textbf{Automated Trajectory Generation (ATG)} framework, an offline data-construction pipeline designed to synthesize and filter interaction trajectories for GUI agents. The pipeline uses an iterative generation--verification--correction procedure to reduce error propagation and hallucinations in multi-step task data; this data-construction loop is separate from policy optimization and is not an online training loop.

\subsubsubsection{\textbf{Multimodal Context Representation}}

At each discrete time step $t$, the system constructs a comprehensive multimodal input $\mathcal{X}_t$:
\begin{equation}
    \mathcal{X}_t = \{ \mathcal{T}_{name}, \mathcal{I}_t, \mathcal{H}_{text}, \mathcal{V}_{hist} \}
\end{equation}
where $\mathcal{T}_{name}$ denotes the task identifier, $\mathcal{I}_t$ is the current RGB screenshot, $\mathcal{H}_{text}$ represents the historical dialogue logs, and $\mathcal{V}_{hist} = \{ \mathcal{I}_{t-1}, \mathcal{I}_{t-2} \}$ provides the temporal visual context through the two preceding frames.

\subsubsubsection{\textbf{Decision Engine: Qwen3-VL-32B}}

\begin{figure*}[t]
  \centering 
    \includegraphics[width=0.95\textwidth]{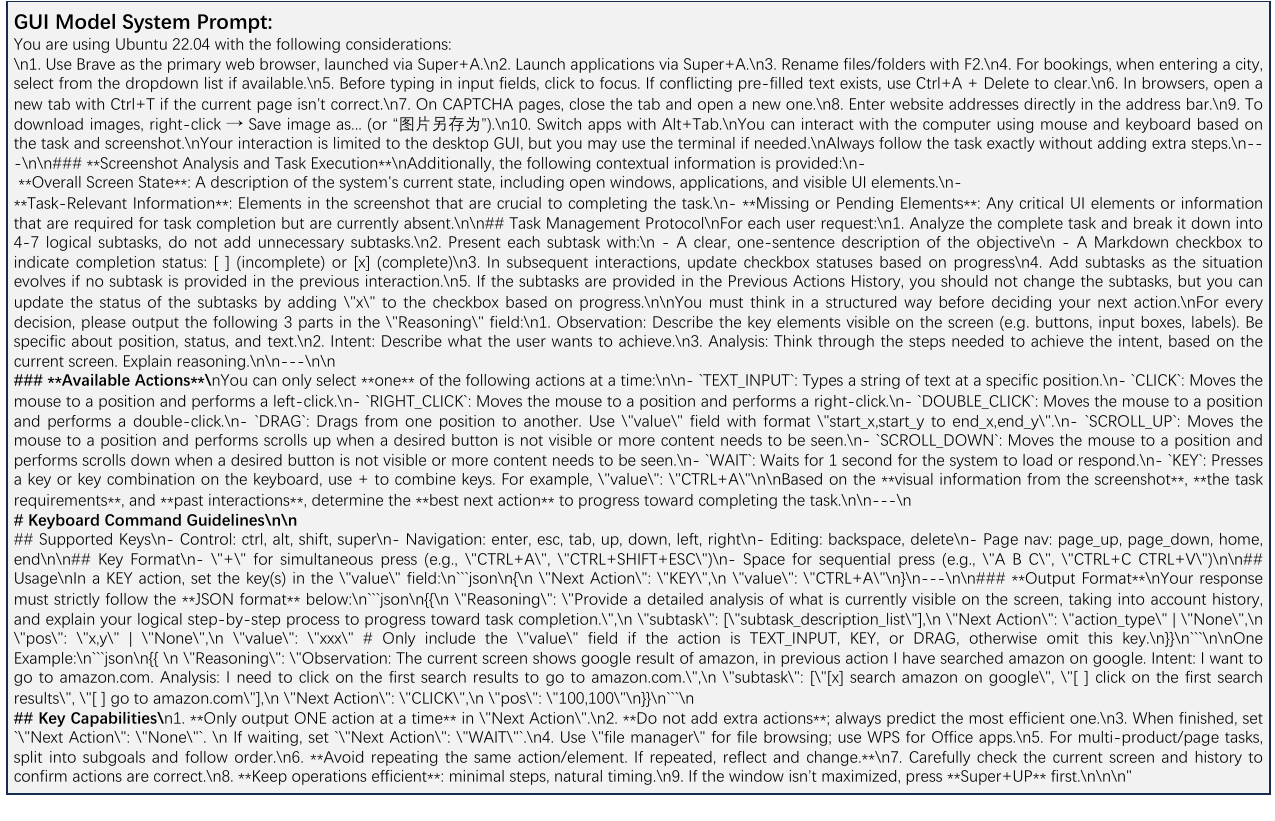}
  \caption{GUI Model System Prompt.}
  \label{fig:gui_agent_prompt}
\end{figure*}

We employ \textbf{Qwen3-VL-32B} as the core GUI Model to parse $\mathcal{X}_t$ and generate a structured decision tuple $\mathcal{D}_t$:
\begin{equation}
    \mathcal{D}_t = \langle \text{Reasoning, Subtasks, Action, Value, Position} \rangle
\end{equation}
The model explicitly outputs its cognitive chain via \textit{Reasoning} and decomposes complex objectives into granular \textit{Subtasks}, ensuring the logical consistency of the resulting \textit{Action} and its corresponding coordinates (\textit{Position}), for reference, see Fig.~\ref{app:format} and system prompt refers to Fig.~\ref{fig:gui_agent_prompt}.

\subsubsubsection{\textbf{Robust Verification and Feedback-driven Refinement}}

\begin{figure*}[t]
  \centering 
    \includegraphics[width=0.95\textwidth]{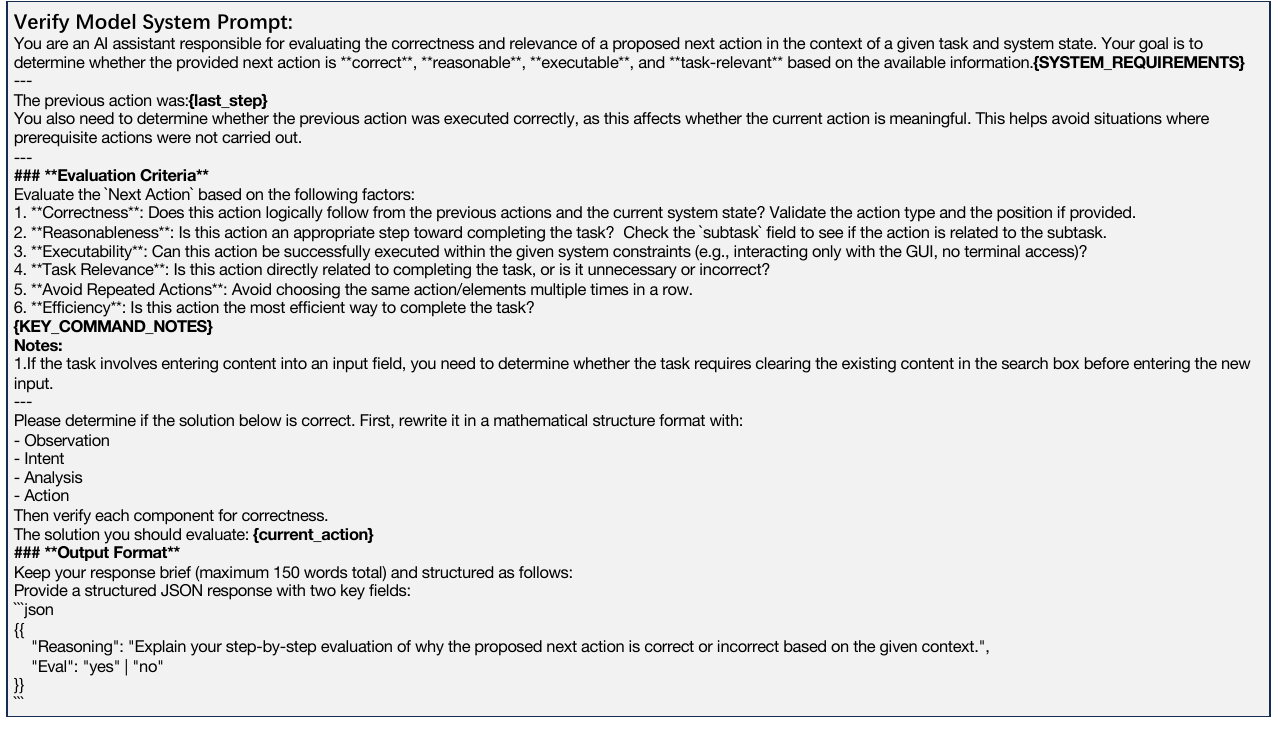}
  \caption{Verify Model System Prompt.}
  \label{fig:verify_prompt}
\end{figure*}

To ensure the reliability of generated actions, a peer \textbf{Qwen3-VL-32B} model acts as the \textbf{Verify Model}, for reference, see Fig.~\ref{app:format} and system prompt refers to Fig.~\ref{fig:verify_prompt}.

\subsubsubsection{\textbf{Majority Voting Mechanism}}

For each candidate action $\mathcal{D}_t$, the Verify Model performs three independent evaluations considering task relevance and planning accuracy. An action is committed to the execution environment (PC) if and only if it secures a majority consensus:
\begin{equation}
    Count(\text{Eval} = \text{"Yes"}) \ge 2
\end{equation}

\subsubsubsection{\textbf{Iterative Self-Correction and Recency-based Fallback}}

If the consensus threshold is not met, the framework initiates a feedback loop (up to $N=3$ iterations):
\begin{itemize}
    \item \textbf{Feedback Injection:} The \textit{Eval} and \textit{Reasoning} from the Verify Model are fed back into the GUI Model as negative constraints to guide policy refinement.
    \item \textbf{Evolutionary Selection:} In cases where the loop reaches the maximum iteration limit without a consensus, the system adopts a \textbf{Recency-based Selection} strategy. It executes the action from the latest iteration that yielded the highest number of "Yes" votes.
    \item \textbf{Rationale:} This strategy assumes that the sequential accumulation of feedback cues progressively narrows the search space, rendering the latest refined output the most robust approximation of the optimal policy.
\end{itemize}

\subsection{\textbf{Data Aggregation and Policy Enhancement}}

Post-execution, successful trajectories are archived into a centralized \textbf{Data Pool}. By integrating these synthesized trajectories with existing open-source datasets, we provide a rich, multi-domain corpus for subsequent post-training stages, including OPD and logged-state policy optimization, thereby continuously improving the agent's generalization capabilities across diverse GUI environments.

\subsubsubsection{\textbf{Guided On-policy Distillation and Policy-Optimization Data Construction}}

The Guided On-policy Distillation and Policy-Optimization Data Construction phase (as illustrated in Figure~\ref{fig:acc}) serves as the core refinement engine for transforming linear explorations into high-density, multi-path expert demonstrations. This process shifts the dataset from a singular-path trajectory to a topological decision graph through state-space exploration and human-in-the-loop auditing.
\begin{itemize}
    \item \textbf{State-space Exploration and Stochastic Inference:} 
    Building upon the multi-step trajectories collected during the ATG phase, the framework performs an in-depth exploration of the state space. For any intermediate state $s_t$ within a trajectory, we employ \textbf{Qwen3-VL-32B} to perform multiple independent stochastic inferences. The objective is to probe the existence of alternative valid solutions for the current sub-task. By varying the decoding parameters, the model generates a set of potential candidate actions $\{a_t^1, a_t^2, \dots, a_t^n\}$, effectively uncovering the inherent strategy diversity of the GUI environment.

    \item \textbf{Expert Auditing and Policy Refinement:} 
    The candidate actions generated during the exploration phase are mapped into a structured semantic format. These candidates are then submitted to a \textbf{Manual Annotation} process, which acts as the ultimate "quality gate." Expert annotators evaluate each proposed action against the visual state and task objectives. As depicted in Figure~\ref{fig:acc}, correct actions that logically advance the task are marked as "True" and preserved, while erroneous or redundant actions are pruned. This human-in-the-loop verification eliminates model hallucinations and ensures the high fidelity of the branched trajectories.

    \item \textbf{Policy-Optimization Data Densification:} 
    The culmination of this process is a structured "Multi-solution" dataset where a single task at a specific step possesses multiple validated paths. This construction significantly densifies the reward landscape for the subsequent \textbf{Guided On-policy Distillation} and \textbf{logged-state policy-optimization} stages. Rather than simple imitation learning, the agent is provided with a rich ensemble of successful strategies, fostering superior decision resilience and error-recovery capabilities.
\end{itemize}

\section{Lite-Dataset}
\label{app:datasets}
\subsection{Task Domains and Environment Configuration}
The \textbf{Lite-Dataset} is a large-scale, high-fidelity multimodal dataset specifically designed for the training and evaluation of autonomous GUI agents. The dataset is strictly rooted in a native PC terminal environment similar to Ubuntu 22.04 LTS, simulating authentic productivity workflows.
\begin{itemize}
    \item \textbf{Full-stack Productivity Ecosystem:} The dataset covers low-level system management (File System, Terminal commands, and System Settings) as well as high-level professional applications, including a full desktop office productivity suite (word processing, spreadsheets, and presentations).
    \item \textbf{Challenge Benchmarks:} We have introduced complex interaction tasks involving VS Code, enterprise instant-messaging clients, and specialized interfaces for virtual-avatar control panels and AI assistants. These tasks are defined as "Challenge Benchmarks," where the decision space exhibits non-linear growth as task complexity increases.
\end{itemize}

\subsection{Standardized Action Space}
To bridge the gap between high-level visual perception and system-level execution, we define a standardized action space $\mathcal{A}$ within the Lite-Datasets. Each action consists of a type, target coordinates (Position), and optional parameters (Value):
\begin{itemize}
    \item \textbf{Mouse Interactions:} Includes \texttt{CLICK}, \texttt{RIGHT\_CLICK}, \texttt{DOUBLE\_CLICK}, \texttt{DRAG}, and \texttt{SCROLL\_UP/DOWN} for page navigation and content scrolling.
    \item \textbf{Keyboard Operations:} Includes \texttt{TEXT\_INPUT} for string entry and \texttt{KEY} for discrete key presses or complex combinations (e.g., \textit{CTRL+A}).
    \item \textbf{System Control:} The \texttt{WAIT} action suspends the agent for 1 second, ensuring system stability and accounting for asynchronous loading latencies.
\end{itemize}

\subsection{Guided On-policy Distillation and Policy-Optimization Pipeline}
\begin{figure*}[t]
  \centering 
    \includegraphics[width=0.95\textwidth]{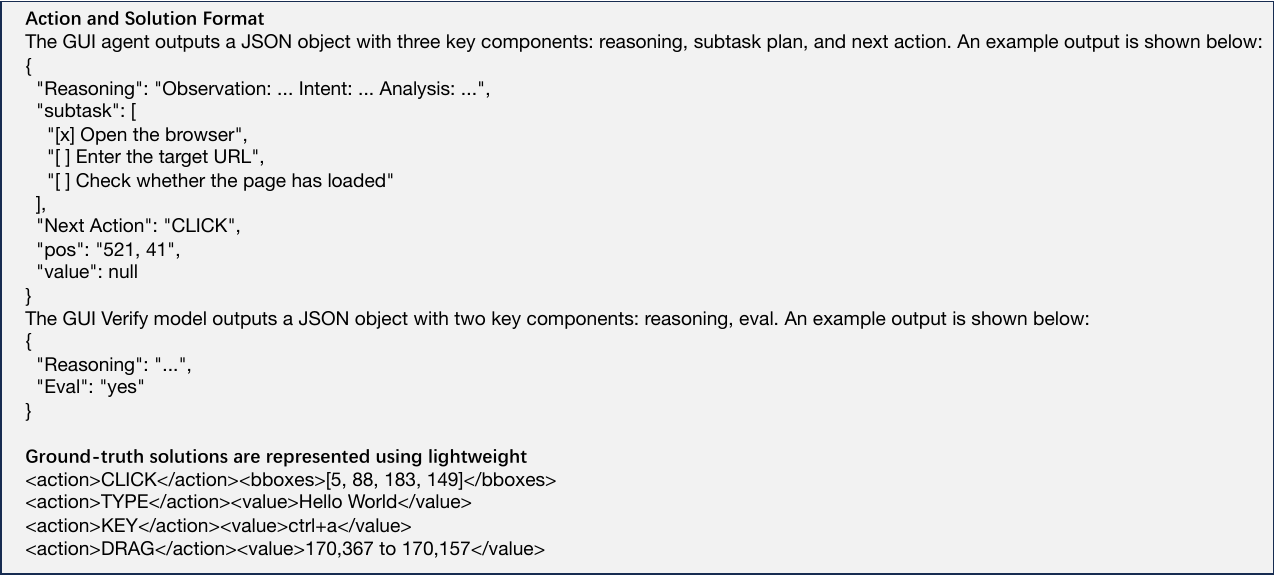}
  \caption{Action and Solution Format And Ground-truth solutions.}
  \label{app:format}
\end{figure*}

To construct the logged training set used by our policy optimization stage, we designed a data construction pipeline based on stochastic exploration. The resulting candidate outputs are stored and scored at their associated GUI states; they are not executed online as part of the optimization loop:
\begin{itemize}
    \item \textbf{Contextual Sliding Window:} Long-horizon trajectories are truncated into manageable samples with a maximum window of three steps (comprising three dialogue turns and three screenshots) to enhance the precision of local decision-making.
    \item \textbf{Multi-solution Stochastic Exploration:} We utilize \textbf{Qwen3-VL-32B} to perform re-inference on the final round of each windowed sample. By increasing the temperature ($Temperature=1.0$), the model performs multiple samplings ($N=5$) to obtain diverse results, which are subsequently deduplicated and stored. The specific output format of these samples follows the template defined in Fig.~\ref{app:format}.
    \item \textbf{Human-in-the-loop Verification:} All candidate paths generated during exploration undergo rigorous manual annotation. We constructed \textbf{30,000} complete multi-step trajectories and approximately \textbf{70,000} derived windowed training samples; across the collected data, \textbf{11,000} samples carry human-verified multi-solution annotations, and the controlled ablations use a fixed \textbf{3,000}-sample annotated subset. 
    Detailed examples of these multi-path ground truths can be found in Fig.~\ref{app:format}.
\end{itemize}

\subsection{Error-aware Trajectory Modeling}
A distinguishing feature of Lite-Dataset is its inclusion of negative supervision signals. Within the complete trajectories collected via the Automated Trajectory Generation (ATG) phase, we preserved and labeled samples where human annotators identified errors in intermediate steps. These labels provide local negative feedback during offline optimization; they should not be interpreted as evidence that the policy was trained with online error recovery in an interactive environment.

\section{Lite-Bench}
\label{app:litebench}
\subsection{Benchmark Overview and Design Philosophy}
To evaluate multi-step GUI task completion, we present \textbf{Lite-Bench}. Built upon a live \textbf{Ubuntu 22.04} environment, Lite-Bench evaluates the policy through \textbf{dynamic, closed-loop interaction}, while training remains based on logged states. The benchmark measures visual grounding, instruction following, multi-step execution, and robustness to the evaluated interface changes; it is not intended to establish online reinforcement-learning or general error-recovery capability.

\subsection{Task Composition and Procedural Constraints}
Lite-Bench comprises 160 tasks, featuring 75 web tasks, 38 terminal tasks, and 47 file system tasks that strictly prohibit terminal usage to enforce pure vision-based interaction. None of the 160 Lite-Bench evaluation task instances is included in the training samples. However, some evaluation tasks share workflow families with training examples while differing in their target entities or object values. We therefore interpret Lite-Bench as an evaluation of transfer to held-out task instances, not as a task-family-disjoint test of generalization to entirely unseen workflows. To simulate the efficiency constraints of professional workflows, each task is capped at a maximum of \textbf{18 steps}. The tasks are categorized into three primary domains:
\begin{itemize}
    \item \textbf{File System:} Focuses on GUI-based file management, directory structuring, and visual data organization. \textit{Notably, Lite-Bench enforces a strict procedural constraint: file system tasks must be accomplished through the graphical file manager. Operations executed via the Terminal for these specific tasks are recorded as failures, ensuring the benchmark measures GUI manipulation proficiency rather than command-line scripting.}
    \item \textbf{Web Browser:} Involves multi-tab navigation, semantic information retrieval, and complex web-form processing.
    \item \textbf{Terminal:} Covers standard command-line operations, system-level configurations, and environment debugging.
\end{itemize}
For a complete evaluation of this fixed task set, the pooled success rate must
satisfy $\mathrm{All}=(47\,\mathrm{File}+75\,\mathrm{Web}+38\,\mathrm{Terminal})/160$,
up to displayed rounding. The same identity applies to an average over
repeated evaluations when every run uses this task set and denominator.
As stated above, OS-World and Lite-Bench results are arithmetic
means over five independent runs. 

\subsection{Evaluation Protocol: LLM-as-a-Judge}
The non-trivial nature of verifying GUI task success requires a nuanced understanding of both final states and interaction sequences. Lite-Bench adopts an automated \textbf{"LLM-as-a-Judge"} protocol utilizing the latest \textbf{GPT-5.4} multimodal model:
\begin{enumerate}
    \item \textbf{Real-time Execution:} The agent interacts with the live Ubuntu environment. Upon task completion or reaching the 18-step limit, the system automatically exports a comprehensive execution log (JSON) and a full-process trajectory of screenshots.
    \item \textbf{Multimodal Auditing:} The execution artifacts are fed into \textbf{GPT-5.4}. The judge model is instructed to perform a dual-criteria audit:
    \begin{itemize}
        \item \textbf{Outcome Verification:} Does the final UI state satisfy the initial user instruction?
        \item \textbf{Compliance Auditing:} Did the agent adhere to the required interaction modality (e.g., using GUI instead of Terminal for file tasks)?
    \end{itemize}
    \item \textbf{Success Rate (SR):} The primary metric is the Success Rate, defined as the percentage of tasks verified as "Successful" by the judge model based on the criteria above.
\end{enumerate}
By combining a live execution environment with a high-level multimodal auditing mechanism, Lite-Bench provides an objective, scalable, and challenging framework for assessing the next generation of general-purpose GUI agents.

Separately from the training-time planning judge evaluated in
Appendix~\ref{app:judge_prompt}, we calibrate the GPT-5.4 benchmark evaluator.
On a held-out human-annotated subset, GPT-5.4 agrees with human judgments at
93.7\% for file tasks, 93.4\% for browser tasks, and 94.8\% for terminal
tasks, with 93.9\% overall agreement.
ScreenSpot-Pro is evaluated over its fixed benchmark images. For OS-World and
Lite-Bench, every reported result for our models and ablations is the
arithmetic mean of five independent runs; the tables do not mix single-run
and repeated-run values. The repetition protocol of literature baselines is
not established here.

To audit train--test independence, we compared Lite-Bench instructions with
approximately 70K training samples using BGE-M3 embeddings and a similarity
threshold of 0.80. Only 862 training samples exceeded the threshold, and no
identical task instructions were found. Lite-Bench covers File System, Web,
and Terminal, whereas Lite-Dataset spans more than seven application
categories, including enterprise collaboration suites, AI assistants, and office productivity tools.

\begin{table}[t]
\centering
\scriptsize
\caption{
Analysis of Guided-OPD.
(a) Teacher supervision quality under privileged guidance.
(b) Student valid-action reliability and solution coverage under repeated sampling.
}
\label{tab:guided_opd_diagnostics}

\resizebox{\columnwidth}{!}{%
\begin{tabular}{lccccc}
\toprule
\multicolumn{6}{c}{\textbf{(a) Teacher Supervision Quality}} \\
\midrule
Method
& Selected Valid $\uparrow$
& Other Valid $\uparrow$
& Invalid $\downarrow$
& V--I Margin $\uparrow$
& Pairwise Acc. $\uparrow$ \\
\midrule
OPD
& 0.596
& 0.314
& 0.000
& 0.455
& 46\% \\
Guided-OPD
& \textbf{0.980}
& \textbf{0.532}
& 0.000
& \textbf{0.756}
& \textbf{77\%} \\
\bottomrule
\end{tabular}%
}

\vspace{0.6em}

\resizebox{\columnwidth}{!}{%
\begin{tabular}{lccc}
\toprule
\multicolumn{4}{c}{\textbf{(b) Student Solution Coverage}} \\
\midrule
Model
& ValidRate@8 $\uparrow$
& UniqueValid@8 $\uparrow$
& Coverage@8 $\uparrow$ \\
\midrule
OPD
& 0.5175
& 0.52
& 0.26 \\
Most-Matched-GT Guided-OPD
& 0.5725
& \textbf{0.66}
& \textbf{0.33} \\
Final MD-GRPO
& \textbf{0.6375}
& \textbf{0.66}
& \textbf{0.33} \\
\bottomrule
\end{tabular}%
}

\end{table}

\section{Analysis of Guided On-Policy Distillation}
\label{app:guided_opd_analysis}
Unless otherwise specified, all diagnostics in this section are conducted on
$N=50$ multi-solution validation states. For the student solution-coverage
analysis, we sample $K=8$ independent completions per state.

We conduct two complementary diagnostic analyses to better understand the effect of
privileged guidance in Guided-OPD. The first evaluates whether privileged guidance
improves the teacher's ability to discriminate valid GUI actions from invalid ones.
The second examines whether Most-Matched-GT causes the student policy to collapse
toward a narrower subset of valid solutions. The results are summarized in
Table~\ref{tab:guided_opd_diagnostics}.

\paragraph{Teacher-scoring diagnostic.}
For each validation state $i$, let $a_i^{\mathrm{sel}}$ denote the human-verified
valid action selected as privileged guidance, $a_i^{\mathrm{oth}}$ denote another
human-verified valid action that is not provided as guidance, and
$a_i^{\mathrm{neg}}$ denote an invalid action sampled from another GUI state.
Let
\[
s_i(a) \in [0,1]
\]
denote the score assigned by the same Qwen3-VL-32B teacher to candidate action $a$
under the corresponding teacher context.

We report five metrics to characterize teacher supervision quality.

\textbf{Selected Valid.}
This metric measures the average teacher score assigned to the valid action selected
as privileged guidance:
\[
S_{\mathrm{sel}}
=
\frac{1}{N}
\sum_{i=1}^{N}
s_i(a_i^{\mathrm{sel}}).
\]
A larger value indicates that the teacher assigns higher confidence to the
guidance-selected valid action.

\textbf{Other Valid.}
To determine whether privileged guidance improves supervision beyond the explicitly
provided reference, we also evaluate another human-verified valid action that is not
used as guidance:
\[
S_{\mathrm{oth}}
=
\frac{1}{N}
\sum_{i=1}^{N}
s_i(a_i^{\mathrm{oth}}).
\]
An increase in this metric indicates that the effect of privileged guidance is not
restricted to the selected reference action, but also improves the teacher's
recognition of alternative valid GUI decisions.

\textbf{Invalid.}
We measure the average score assigned to an invalid action:
\[
S_{\mathrm{neg}}
=
\frac{1}{N}
\sum_{i=1}^{N}
s_i(a_i^{\mathrm{neg}}).
\]
A lower value is preferred, as it indicates that the teacher does not assign high
confidence to actions that are invalid for the current GUI state.

\textbf{Valid--Invalid Margin.}
We measure the separation between valid and invalid actions by averaging the scores
of the two valid candidates and subtracting the invalid-action score:
\[
M_{\mathrm{V-I}}
=
\frac{S_{\mathrm{sel}} + S_{\mathrm{oth}}}{2}
-
S_{\mathrm{neg}}.
\]
A larger margin indicates that the teacher more clearly separates valid GUI actions
from invalid alternatives, yielding a more discriminative supervision signal.

\textbf{Pairwise Preference Accuracy.}
We further measure whether the teacher ranks valid actions above the invalid action.
For each state, we compare both valid candidates against the invalid candidate:
\begin{equation*}
\resizebox{\columnwidth}{!}{$
\mathrm{PairAcc}
=
\frac{1}{2N}
\sum_{i=1}^{N}
\left[
\mathbb{I}\!\left(
s_i(a_i^{\mathrm{sel}}) > s_i(a_i^{\mathrm{neg}})
\right)
+
\mathbb{I}\!\left(
s_i(a_i^{\mathrm{oth}}) > s_i(a_i^{\mathrm{neg}})
\right)
\right],
$}
\end{equation*}
where $\mathbb{I}[\cdot]$ is the indicator function.
A larger value means that the teacher more consistently prefers valid actions over
invalid ones.

As shown in Table~\ref{tab:guided_opd_diagnostics}(a), Guided-OPD increases the
Selected Valid score from 0.596 to 0.980 and the Other Valid score from 0.314 to
0.532, while the Invalid score remains at 0.000. Accordingly, the Valid--Invalid
Margin increases from 0.455 to 0.756, and Pairwise Preference Accuracy improves
from 46\% to 77\%. Importantly, the improvement on Other Valid indicates that
privileged guidance does more than simply increase confidence in the provided
reference action; it also improves the teacher's discrimination of alternative valid
GUI decisions.

\paragraph{Student solution-coverage diagnostic.}
We next examine whether the stronger alignment introduced by Most-Matched-GT reduces
the diversity of valid actions produced by the student. For each validation state
$i$, we independently sample $K=8$ completions from the student policy:
\[
\hat{a}_{i,1}, \hat{a}_{i,2}, \ldots, \hat{a}_{i,K}.
\]
Let $A_i^{*}$ denote the human-verified valid-action set for state $i$.
A sampled action is considered valid if it matches at least one action in $A_i^{*}$
according to the GUI action-matching function. We define
\[
m_{i,j}
=
\mathbb{I}
\left[
\exists a^{*} \in A_i^{*},
\;
\hat{a}_{i,j} \sim a^{*}
\right],
\]
where $\hat{a}_{i,j} \sim a^{*}$ denotes that the sampled action matches the
corresponding human-verified valid action.

We report three metrics to quantify valid-action reliability and solution diversity.

\textbf{ValidRate@8.}
ValidRate@8 measures the fraction of the eight sampled actions that are valid:
\[
\mathrm{ValidRate@8}
=
\frac{1}{N}
\sum_{i=1}^{N}
\frac{1}{K}
\sum_{j=1}^{K}
m_{i,j},
\qquad K=8.
\]
This metric measures the reliability of the student policy under repeated sampling:
a larger value indicates that a greater fraction of generated actions are valid.

\textbf{UniqueValid@8.}
For each state, we define the set of distinct human-verified valid actions recovered
by the sampled outputs as
\[
U_i
=
\left\{
a^{*}\in A_i^{*}
\;\middle|\;
\exists j \in \{1,\ldots,K\},
\hat{a}_{i,j} \sim a^{*}
\right\}.
\]
We then compute
\[
\mathrm{UniqueValid@8}
=
\frac{1}{N}
\sum_{i=1}^{N}
|U_i|.
\]
Unlike ValidRate@8, which can be high even when the same valid action is repeatedly
generated, UniqueValid@8 measures how many distinct valid action modes are observed
under repeated sampling.

\textbf{Coverage@8.}
Coverage@8 measures the fraction of the annotated valid-action set recovered by the
student samples:
\[
\mathrm{Coverage@8}
=
\frac{1}{N}
\sum_{i=1}^{N}
\frac{|U_i|}{|A_i^{*}|}.
\]
A larger value indicates that the student covers a larger fraction of the
human-verified valid solution space.

As shown in Table~\ref{tab:guided_opd_diagnostics}(b), Most-Matched-GT Guided-OPD
improves ValidRate@8 from 0.5175 to 0.5725, while increasing UniqueValid@8 from
0.52 to 0.66 and Coverage@8 from 0.26 to 0.33. The final MD-GRPO model further
increases ValidRate@8 to 0.6375 while preserving UniqueValid@8 and Coverage@8 at
0.66 and 0.33, respectively. These results indicate that Most-Matched-GT improves
the reliability of valid-action generation without reducing the observed diversity
or coverage of valid solutions.

\section{Inference Efficiency and Training Cost}
\label{app:efficiency}

We benchmarked inference with 
three-window GUI contexts. On one
A100, Qwen3-VL-2B reaches 103.0 tokens/s at approximately 4.5\,GB VRAM,
LiteGUI-30B-A3B reaches 91.54 tokens/s at approximately 18\,GB, whereas the
Qwen3-VL-32B teacher reaches 27.27 tokens/s and requires two A100 GPUs with
tensor parallelism. On one RTX 4090, Qwen3-VL-2B reaches 81.2 tokens/s in
BF16 and LiteGUI-30B-A3B reaches 17.7 tokens/s with AWQ; the dense 32B AWQ
model runs out of memory.

The 11K multi-solution annotations required approximately 120 human hours.
The full data-generation and training pipeline required approximately 1.2K
A100 GPU hours. These are one-time offline costs; they trade against lower
inference memory and latency for the resulting lightweight models.

Under the same controlled training configuration, the SFT baseline requires
approximately 4 GPU-hours, compared with 17 GPU-hours for
Most-Matched-GT-Guided OPD. In an additional ShowUI-Desktop-8K comparison with
identical model settings, SFT, Most-Matched-GT-Guided OPD, and GRPO require
approximately 0.5, 4, and 12 hours, respectively, and achieve 30.55\%,
46.29\%, and 43.51\% ScreenSpot-Pro accuracy.

\section{Limitations and Broader Impacts}
\label{app:limitations}
Although our method improves GUI-agent training through guided on-policy distillation and multi-solution dual-level optimization on logged states, several limitations remain. The GRPO terminology refers to the optimization algorithm; in our implementation, candidate GUI actions are scored without executing them in the environment.

First, the human-verified valid action set $\mathcal{A}_t^*$ is only a finite approximation of the full valid action space. GUI tasks often admit many semantically equivalent action paths, and our annotations cannot exhaustively cover all possible correct actions. As a result, a model output that correctly advances the task may still receive a lower reward if the corresponding action is not included in the annotated multi-solution set. The Most-Matched-GT selector should therefore be understood as an alignment heuristic that reduces supervision--state mismatch, not as a guarantee that all valid action modes remain equally represented.

Second, constructing multi-solution annotations requires additional human verification. Compared with single-action supervision, annotating and validating multiple acceptable actions for the same GUI state is more expensive and may limit scalability to new applications, operating systems, or task domains.

Third, the planning reward relies on a frozen Qwen3-VL-32B-Instruct model as an external VLM judge during the logged-state policy-optimization stage. Although the judge is not used at inference time, it increases training cost and may introduce model-specific evaluation bias. In particular, the judge may over- or under-estimate the quality of a subtask plan when the screenshot history is ambiguous or when the plan uses a valid but uncommon strategy.

Fourth, GRPO is a practical choice for our logged-state setting rather than a claim of superiority over critic-based PPO. PPO may support more fine-grained state-to-state reward modeling when reliable value targets and online rollouts are available, while our setting avoids an additional critic and the cost of repeated GUI interaction. A compute-matched GRPO--PPO comparison remains future work.

Fifth, our latency and memory measurements are server-class measurements and do not include a full energy study or a broad on-device deployment evaluation. Our experiments focus on GUI-agent tasks under the evaluated benchmark and data distribution. The results may not directly generalize to all operating systems, applications, screen resolutions, languages, or highly dynamic interfaces. Further evaluation is needed before deploying such agents in open-ended real-world environments.

Finally, GUI automation systems can have both positive and negative societal impacts. They may improve productivity and accessibility by automating repetitive computer tasks, but they may also cause unintended operations, privacy leakage from screenshots, or misuse in unauthorized automation. Practical deployment should include permission control, sensitive-action confirmation, logging, and privacy filtering for screenshot-based data.

\section{Existing Assets and Licenses}
\label{app:assets}

We use several existing models, benchmarks, and software packages in this work. We cite the corresponding papers or official repositories when applicable and follow their license and usage terms. Table~\ref{tab:existing_assets} summarizes the main existing assets used in our experiments.

\begin{table}[t]
\centering
\small
\caption{Existing assets used in this work.}
\label{tab:existing_assets}
\resizebox{\linewidth}{!}{%
\begin{tabular}{p{0.25\linewidth} p{0.55\linewidth} p{0.15\linewidth}}
\toprule
Asset & Role in this work & License / Terms \\
\midrule
Qwen3-VL-32B~\citep{Qwen3-VL}
& Frozen VLM judge for the planning reward; also used as the teacher model for Guided OPD when applicable.
& Apache-2.0 \\

Qwen3-VL-2B~\citep{Qwen3-VL}
& Base lightweight student model for GUI-agent training and controlled ablations.
& Apache-2.0 \\

Qwen3-VL-30B-A3B~\citep{Qwen3-VL}
& Base student model for GUI-agent training.
& Apache-2.0 \\

OSWorld~\citep{OSWorld}
& Public benchmark for evaluating GUI agents in real computer environments.
& Apache-2.0 \\

ScreenSpot-Pro~\citep{li2025screenspotpro}
& Public GUI grounding benchmark for professional high-resolution computer-use scenarios.
& MIT \\

ms-swift 
& Main training framework for SFT, GKD/OPD, and GRPO experiments.
& Apache-2.0 \\

Megatron-LM / Megatron Core 
& Distributed training backend used through the Megatron training path, including tensor, pipeline, data, and expert parallelism.
& Apache-2.0 \\
\bottomrule
\end{tabular}
}%
\end{table}

For all existing assets, we use them only for research purposes and follow their official license and usage terms. We do not redistribute third-party pretrained models or benchmark assets with restrictive terms; instead, we provide instructions for obtaining them from their official sources when necessary. The released assets include our own GUI trajectories, annotations, prompts, configurations, and code components that we are permitted to distribute.

\section{Qualitative Visual Case Studies}
\label{app:qualitative}

To complement the quantitative benchmarks, Figure~\ref{fig:qualitative_main} presents real desktop interaction examples from both offline multi-solution training data and live closed-loop OSWorld evaluation.

\paragraph{Multi-modal equivalence in logged GUI states.}
As shown in Figure~\ref{fig:qualitative_main}(a--b), standard desktop interfaces naturally expose multiple valid interaction affordances at the same visual state: opening a context menu via \texttt{RIGHT\_CLICK} on blank workspace ($a^{*(1)}$), clicking the top-bar \texttt{`...'} icon ($a^{*(2)}$), or pressing \texttt{Ctrl+Shift+N} ($a^{*(3)}$). Under single-reference SFT or single-GT RL, sampling $a^{*(2)}$ when the dataset logged $a^{*(3)}$ incurs a false-negative penalty. By matching against the verified set $\mathcal{A}_t^*$, Most-Matched Guided-OPD supplies a student-aligned privileged reference to the teacher, while MD-GRPO rewards any valid mode.

\paragraph{Preventing repetitive click loops and state hallucination in OSWorld.}
Figure~\ref{fig:qualitative_main}(c--d) compares closed-loop rollouts of Qwen3-VL-2B + SFT and LiteGUI-2B on OSWorld (\texttt{libreoffice\_writer}). SFT prematurely marks \texttt{[x] Select all text} without highlighting the body text and falls into a 15-step repetitive click loop on the font dropdown (\texttt{pos=(370,144)}). In contrast, LiteGUI-2B grounds its subtask status in visual evidence (the highlighted text body and \texttt{Selected: 736 words} indicator on the status bar), executes \texttt{Ctrl+A} before editing the font box, and completes the task in 5 steps.

\section{Closed-Loop Trajectory and Failure Audit on OSWorld}
\label{app:osworld_trajectory_audit}

To examine how offline optimization on logged GUI states behaves across different interaction horizons and failure modes during live closed-loop execution, we conduct a fine-grained turn-by-turn audit on a fully logged single-run OSWorld evaluation (361 tasks, 15-step cap, screenshot-only observations) of our 30B-A3B offline-optimized model, where every intermediate screenshot, model output, and action record is archived. While Table~\ref{tab:osworld} reports the five-run mean ($22.7\%$), this fully archived single-run audit achieves an overall score of $79.78/361 = 22.10\%$ ($95\%$ bootstrap CI $[18.13\%, 26.53\%]$, with $78$ full-score, $2$ partial-score, and $281$ zero-score tasks). On the tasks with paired trajectories from the off-the-shelf Qwen3-VL-30B-A3B backbone, offline post-training raises the closed-loop score from $16.52\%$ to $22.22\%$, yielding a paired gain of $+5.71$ percentage points ($95\%$ paired bootstrap CI $[+2.18, +9.25]$, one-sided $p=0.0010$), with $32$ tasks flipping from failure to full success versus $10$ regressions.

\begin{table}[t]
\centering
\scriptsize
\caption{\textbf{Closed-loop OSWorld breakdown by executed trajectory length} on paired tasks (15-step cap) between the off-the-shelf 30B-A3B backbone and the fully logged LiteGUI-30B-A3B audit run. ``Full Succ.'' counts episodes achieving a full score of $1.0$; ``$0\!\rightarrow\!1$ Gains'' counts tasks where the backbone scores $0.0$ and LiteGUI-30B-A3B achieves $1.0$.}
\label{tab:supp_osworld_horizon}
\resizebox{\columnwidth}{!}{%
\begin{tabular}{lcccccc}
\toprule
\multirow{2}{*}{Step Horizon} & \multicolumn{2}{c}{Backbone (30B-A3B)} & \multicolumn{2}{c}{LiteGUI-30B-A3B} & \multirow{2}{*}{$\Delta$ Full Succ.} & \multirow{2}{*}{$0\!\rightarrow\!1$ Gains} \\
\cmidrule(lr){2-3} \cmidrule(lr){4-5}
 & Full Succ. & SR (\%) & Full Succ. & SR (\%) & & \\
\midrule
1--4 steps   & 9  & 42.86 & 11 & 68.75 & $+2$  & 3  \\
5--8 steps   & 26 & 66.67 & 31 & 59.19 & $+5$  & 9  \\
9--12 steps  & 10 & 38.46 & 13 & 61.90 & $+3$  & 6  \\
13--15 steps & 11 & 5.35  & \textbf{22} & \textbf{8.58} & \textbf{$+11$} & \textbf{13} \\
\midrule
All paired & 56 & 16.52 & \textbf{77} & \textbf{22.22} & \textbf{$+21$} & \textbf{32} \\
\bottomrule
\end{tabular}%
}
\end{table}

\paragraph{Gains concentrate in 13--15 step trajectories.}
Table~\ref{tab:supp_osworld_horizon} stratifies paired OSWorld episodes by executed trajectory length. Rather than concentrating only on short 1--4 step tasks, the largest net increase occurs in the 13--15 step regime, where full-score completions double from $11$ to $22$ ($+11$) and contribute $13$ of the $32$ newly solved tasks. Moreover, among the $206$ tasks where the backbone exhausts all $15$ steps with zero score, the offline-optimized policy rescues $21$ tasks ($10.2\%$) to full completion across execution lengths ranging from $3$ to $15$ steps.

\paragraph{Structured state tracking and unprompted local recovery.}
Auditing all $4,603$ model turns across $360$ logged sessions shows $100.0\%$ ($4,603/4,603$) valid JSON schema adherence and active subtask checkbox updates (\texttt{[x]}) in $99.2\%$ ($357/360$) of episodes. Even under a strict standalone audit that excludes any episode with external runtime retry hints (isolating $145$ purely unprompted episodes that encounter unexpected visual states), the policy spontaneously recognizes the visual state discrepancy in its \texttt{Reasoning} field and revises its next action, succeeding in $49/145$ ($33.8\%$) tasks. Representative examples include: (i)~Chrome task \texttt{2ad9387a} ($13$ steps, $0.0\!\rightarrow\!1.0$), where Step~1 inadvertently closes the browser window, Step~2 detects from the desktop screenshot that Chrome is no longer open and inserts a relaunch subtask from the dock, and the remaining steps complete creation of the \texttt{Favorites} bookmark folder; and (ii)~VS Code task \texttt{0512bb38} ($11$ steps, $0.0\!\rightarrow\!1.0$), where clicking a three-dot menu at Step~4 exposes editor options rather than extension options, prompting the agent to pivot to the Command Palette (\texttt{Ctrl+Shift+P}) to install the local \texttt{.vsix} package.

\begin{table}[t]
\centering
\scriptsize
\caption{\textbf{Mutually exclusive failure taxonomy on OSWorld} across all 281 zero-score tasks in the 361-task trajectory audit.}
\label{tab:supp_osworld_failures}
\resizebox{\columnwidth}{!}{%
\begin{tabular}{lcc}
\toprule
Failure Mode & Count & Share of Zero-Score Tasks (\%) \\
\midrule
Horizon exhausted (15 steps reached, no loop) & 128 & 45.6 \\
Stuck in local action loop ($\ge 3$ identical consecutive actions) & 99 & 35.2 \\
Premature termination (\texttt{DONE} $<15$ steps, incomplete) & 42 & 14.9 \\
Excessive waiting / service latency ($\ge 5$ \texttt{WAIT} actions) & 6 & 2.1 \\
Other early stop or environment error & 6 & 2.1 \\
\midrule
Total zero-score tasks & 281 & 100.0 \\
\bottomrule
\end{tabular}%
}
\end{table}

\paragraph{Failure taxonomy and remaining bottlenecks.}
Table~\ref{tab:supp_osworld_failures} partitions all $281$ zero-score episodes into five mutually exclusive categories. Two bottlenecks account for $80.8\%$ of failures: \emph{15-step horizon exhaustion without looping} ($45.6\%$), where the policy continues exploring plausible UI regions in dense productivity suites (e.g., LibreOffice Calc and multi-app workflows) without locating the required control in time, and \emph{local action loops} ($35.2\%$), where repeated pointer or keyboard operations fail to alter the underlying application state. Another $14.9\%$ arise from premature \texttt{DONE} predictions before a final confirmation or secondary constraint is satisfied. Together, these quantitative diagnostics clarify both how logged-state multi-solution supervision transfers to multi-step execution and why online environment interaction remains an important direction for deeper exploration and terminal verification.

\clearpage

\begin{figure*}[t]
  \centering 
    \includegraphics[width=0.95\textwidth]{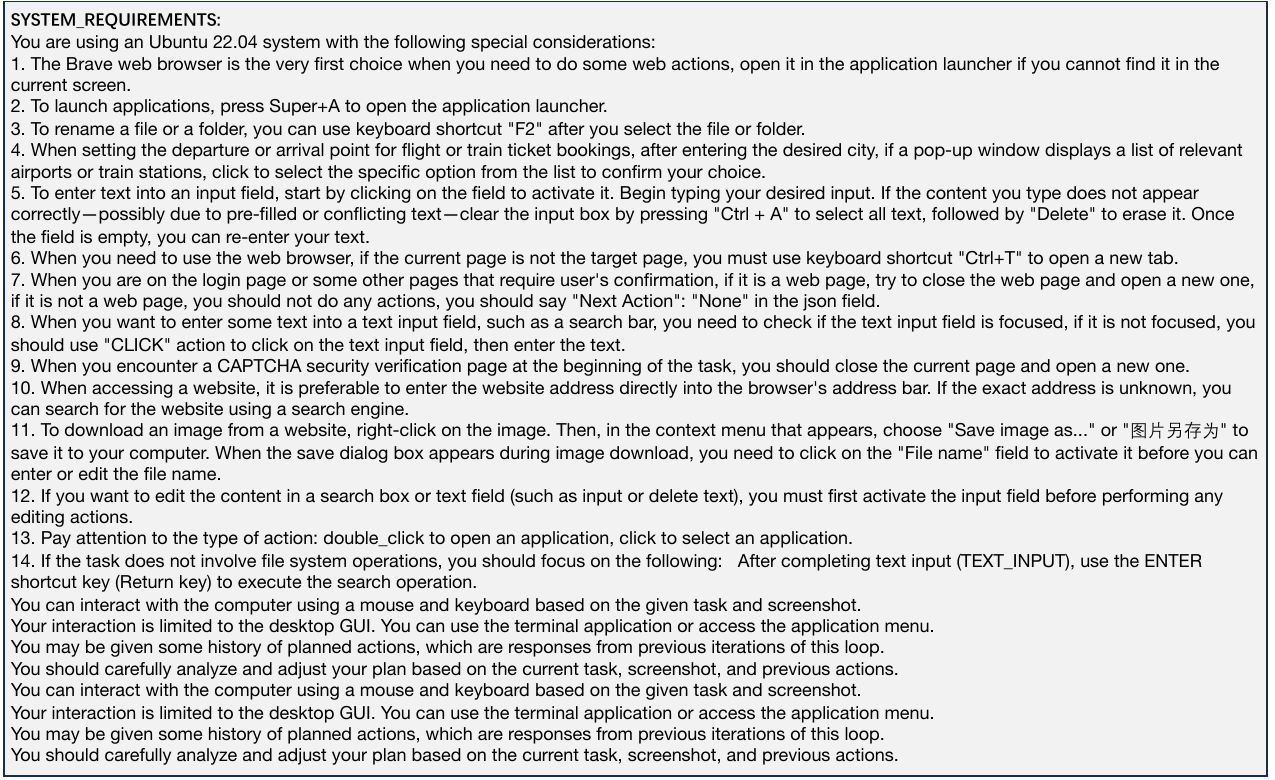}
  \caption{System requirements Prompt.}
  \label{fig:sys_note_prompt}
\end{figure*}
\begin{figure*}[t]
  \centering 
    \includegraphics[width=0.95\textwidth]{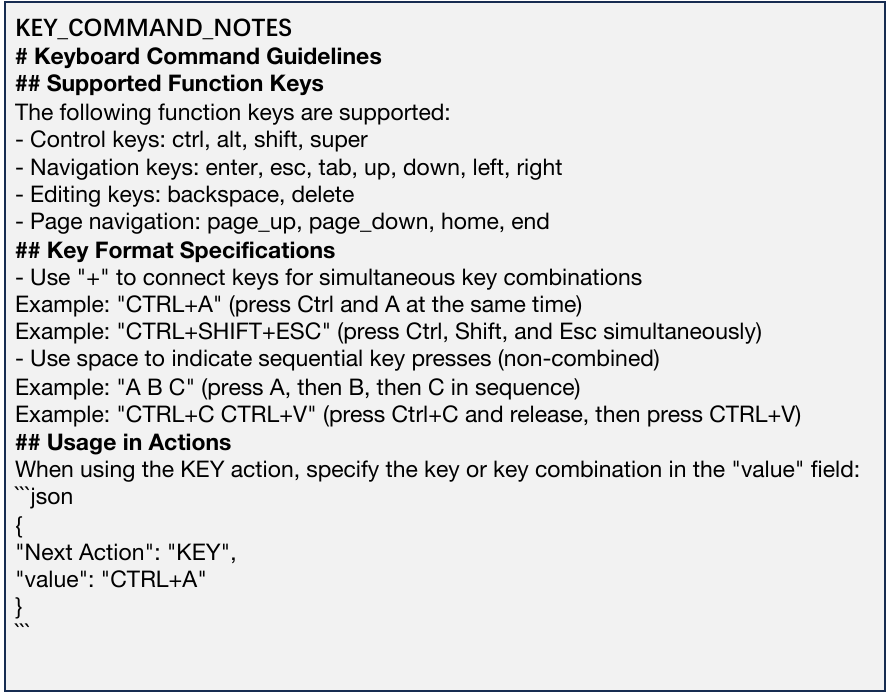}
  \caption{Key command note Prompt.}
  \label{fig:key_prompt}
\end{figure*}

\end{document}